\documentclass[dvipsnames]{article}

\PassOptionsToPackage{nopatch=footnote}{microtype}
\usepackage{colm2024_conference}

\usepackage[utf8]{inputenc}
\usepackage[T1]{fontenc}
\usepackage{hyperref}
\usepackage{url}
\usepackage{booktabs}
\usepackage{amsfonts}
\usepackage{nicefrac}
\usepackage{xcolor}
\usepackage{textcomp}
\usepackage{graphicx}
\usepackage{multirow}
\usepackage{color}
\usepackage{threeparttable}
\usepackage{tablefootnote}
\usepackage{arydshln}
\usepackage{amssymb}
\usepackage{pifont}
\usepackage{wrapfig}
\usepackage{verbatim}
\usepackage{enumitem}
\usepackage{amsmath}
\usepackage{colortbl}
\usepackage{setspace}
\usepackage{hhline}
\usepackage{subcaption}
\usepackage{makecell}
\usepackage{mathrsfs}
\usepackage{tikz}
\usetikzlibrary{arrows.meta,positioning,fit,calc}
\usepackage{tabularx}
\usepackage{fontawesome5}
\usepackage{algorithm}
\usepackage{algorithmic}

\newcommand{\MYBENCH}{\textsc{EvoCode-Bench}}
\newcommand{\equalcontrib}{\textsuperscript{*}}
\newcommand{\corresponding}{\textsuperscript{\ensuremath{\dagger}}}
\newcommand{\projectlead}{\textsuperscript{\ensuremath{\ddagger}}}

\definecolor{PKURed}{HTML}{7B1128}
\definecolor{PKUDarkRed}{HTML}{4B0014}
\definecolor{PKULightRed}{HTML}{F3E7EA}
\definecolor{PKURose}{HTML}{B98590}
\definecolor{PKUPlum}{HTML}{8E5A67}
\definecolor{PKUGray}{HTML}{6E676A}

\newcolumntype{C}[1]{>{\centering\arraybackslash}p{#1}}
\newcolumntype{L}[1]{>{\raggedright\arraybackslash}p{#1}}

\title{\MYBENCH: Evaluating Coding Agents in Multi-Turn Iterative Interactions}

\author{%
  \textbf{Haiyang Shen$^{1,2}$\equalcontrib,\quad Xuanzhong Chen$^{1,3}$\equalcontrib,\quad Wendong Xu$^{1,4}$\equalcontrib\projectlead}\\
  \textbf{Yun Ma$^{2}$\corresponding,\quad Liang Chen$^{1}$\corresponding,\quad Kuan Li$^{1}$\corresponding}\vspace{8pt}\\
  $^{1}$UniPat AI \quad
  $^{2}$Peking University \quad
  $^{3}$Tsinghua University \quad
  $^{4}$HKU \\
  \texttt{\small \{haiyangshen,xuanzhongchen,wendongxu,liangchen,kuanli\}@unipat.ai}, \texttt{\small mayun@pku.edu.cn} \vspace{1pt}\\
  {\small \equalcontrib Equal contribution. \quad \projectlead Coding Project Leader. \quad \corresponding Corresponding authors.}\vspace{8pt}
}

\begin{document}

\maketitle

\begin{abstract}

Coding agents are increasingly used as iterative development partners, but most benchmarks still evaluate one specification followed by one final assessment. This leaves out a basic question: can an agent keep its own codebase working as requirements change? We introduce \MYBENCH, a benchmark of 26 stateful coding tasks and 227 evaluated rounds. Each task preserves the agent's workspace for 5--15 rounds, states requirements through observable behavior, and uses cumulative executable tests to check new requirements and still-active prior ones. We evaluate 13 coding agents with two metrics: MT@4, a four-attempt fail-stop multi-round score, and SR, a single-round score from a reference-completed prior state. For most agents, SR exceeds MT@4 by 22--40 points. The gap also changes rankings: the highest-SR agent (78.9) ranks only third in persistent execution (44.0 MT@4). Even the strongest agents achieve only about 50\% success on multi-turn metrics, and aggregate pass rate drops below half of round-1 performance by round~5. Failure analysis shows tier-dependent behavior: weaker agents fail early, while stronger agents survive long enough to expose specification-tracking and regression failures. We release the benchmark data and Harbor multi-turn infrastructure.

\end{abstract}

\section{Introduction}
\label{sec:Introduction}

Coding agents have evolved from code-completion tools into systems that plan, edit files, execute commands, and interact with development environments. Products such as Claude Code~\cite{anthropic2026claudecode}, Cursor~\cite{cursor2026}, Codex~\cite{openai2026codex}, and Windsurf~\cite{windsurf2026} are now used for tasks such as debugging, data analysis, service deployment, and infrastructure management. In these settings, the agent must operate within a persistent workspace: each edit changes files, dependencies, interfaces, and tests that later interactions inherit. This makes coding a useful setting for evaluating tool-using agents because decisions leave executable traces and correctness can be checked through behavior.

Current evaluation only partly matches this setting. Function-level benchmarks~\cite{chen2021evaluatinglargelanguagemodels,austin2021programsynthesislargelanguage,DBLP:conf/iclr/JainHGLYZWSSS25,DBLP:conf/iclr/ZhuoVCH0WYZHPB025} evaluate localized programming ability through one-shot generation. Repository-level benchmarks test repair, completion, and feature addition in realistic codebases~\cite{DBLP:conf/iclr/0003XM24,DBLP:conf/iclr/JimenezYWYPPN24,DBLP:conf/iclr/YangJZLYWPMSNY025,li2024deveval,lehai2025repoexec,ding2026nl2repobench}. SWE-Gym~\cite{DBLP:conf/icml/Pan0NJ0S025} turns software engineering tasks into training environments. Environment-level benchmarks such as AppWorld~\cite{DBLP:conf/acl/TrivediKHMDLGSB24} and Terminal-Bench~\cite{merrill2026terminalbenchbenchmarkingagentshard} evaluate agents that interact with executable environments. Multi-turn evaluation has also been studied in dialogue, tool use, and general agent settings~\cite{DBLP:conf/nips/ZhengC00WZL0LXZ23,kwan2024mteval,laban2026lost,yao2024taubenchbenchmarktoolagentuserinteraction,DBLP:conf/naacl/LuHZANBMMLYWP25}. These efforts make coding evaluation more realistic, but their usual unit is still one task specification followed by one terminal assessment. Persistent coding adds a different requirement: later turns must build on the agent's own artifact while preserving all active requirements.

In practice, users interact with coding agents iteratively: they add output formats, correct behavior, introduce constraints, and deprecate earlier requirements. Unlike dialogue, where an earlier poor response can often be corrected in text, coding materializes decisions into file layouts, schemas, APIs, and dependency choices. These commitments constrain later work and can create regressions. A benchmark for this setting needs cross-round dependencies, cumulative executable tests, and scoring that identifies when the evolving workspace first stops satisfying the active specification.

To address this gap, we introduce \MYBENCH, a benchmark for coding agents in interactive, persistent multi-turn sessions where requirements evolve and sometimes conflict. \MYBENCH~ contains 26 multi-round tasks and 227 evaluated rounds, with 5 to 15 rounds per task. The benchmark is grouped by two axes: \emph{Interaction style}, which describes how users communicate across rounds, and \emph{Engineering activity}, which describes the kind of code change each round asks for. Since correct agents may build different internal designs, \MYBENCH~ evaluates behavior rather than implementation paths: instructions state observable requirements, and verification scripts exercise those requirements through execution instead of inspecting code structure. Section~\ref{sec:Method} presents the full taxonomy and task construction pipeline.

We evaluate 13 leading coding agents using the Terminus-2 harness~\cite{laude2026terminus2} and Harbor execution protocol~\cite{laude2026harbor}. We extend Harbor so that one Docker workspace and one agent session persist across round boundaries while cumulative verifiers are swapped and state lineage is recorded. The evaluation gives three findings. First, SR exceeds MT@4 by 22 to 40 points for most agents, and the gap reranks models: Opus-4.6 has the highest SR (78.9) but only the third-highest MT@4 (44.0). Second, \MYBENCH~ is challenging even for the strongest models: only two agents exceed half of the multi-round credit, and top agents still fail on many long trajectories. Third, performance degrades rapidly with interaction depth. Failure analysis shows tier-dependent breakdown patterns: lower-tier agents typically miss early requirements, whereas stronger agents struggle with conflict resolution, self-correction, and regression management in later turns. These results point to preserving correctness across evolving requirements as a separate evaluation target.

\paragraph{Contributions.}
We contribute: (i) a task definition for interactive persistent multi-turn coding with evolving and conflicting requirements; (ii) \MYBENCH, with 26 tasks, 227 evaluated rounds, persistent workspaces, reference deltas, cumulative verifiers, fail-stop scoring, and a two-axis taxonomy; (iii) open-source Harbor multi-turn extensions for continuous sessions, verifier swaps, reference fast-forwarding, snapshot/resume lineage, and fail-stop accounting; and (iv) a 13-agent evaluation showing that single-round performance can overstate persistent reliability and change model ranking.

\section{Related Work}
\label{sec:RelatedWork}

\paragraph{Code Generation and Repair Benchmarks}
\label{sec:rw_code_bench}

Code evaluation benchmarks have progressed from function-level generation~\cite{chen2021evaluatinglargelanguagemodels,austin2021programsynthesislargelanguage} through instruction-following with library calls~\cite{DBLP:conf/iclr/JainHGLYZWSSS25,DBLP:conf/iclr/ZhuoVCH0WYZHPB025} to repository-level task solving. SWE-bench~\cite{DBLP:conf/iclr/JimenezYWYPPN24} and SWE-bench Multimodal~\cite{DBLP:conf/iclr/YangJZLYWPMSNY025} require agents to resolve real issues; RepoBench~\cite{DBLP:conf/iclr/0003XM24} evaluates repository-level completion. A parallel line studies repository-level generation: EvoCodeBench~\cite{li2024evocodebench}, DevEval~\cite{li2024deveval}, RepoExec~\cite{lehai2025repoexec}, CodeS~\cite{zan2024codes}, NL2Repo-Bench~\cite{ding2026nl2repobench}, and NoCode-bench~\cite{deng2025nocodebench} each advance different facets of repository construction and feature implementation. SWE-Gym~\cite{DBLP:conf/icml/Pan0NJ0S025} turns software engineering tasks into training environments. Two recent efforts add sequential structure: SWE-EVO~\cite{zhao2025sweevo} requires implementing an entire software release under cumulative regression tests from a single specification, and SlopCodeBench~\cite{garg2025slopcodebench} evaluates code quality across sequential additive checkpoints. While both test sustained correctness within a single session, they rely on fixed or additive specifications without interactive user dialogue or requirement revision. \MYBENCH~ distinguishes itself through an inherently interactive multi-turn design: requirements evolve and conflict across rounds through simulated user dialogue, and the cumulative verifier checks the agent's own accumulated workspace at every round. This explicitly measures regression avoidance, conflict resolution, and incremental adaptation under requirement change that fixed-specification benchmarks do not measure.

\paragraph{Autonomous Coding Agent Evaluation}
\label{sec:rw_agent_eval}

Recent research on autonomous agents explores systems capable of planning, executing shell commands, and interacting with development environments. SWE-agent~\cite{DBLP:conf/nips/YangJWLYNP24} and OpenHands~\cite{DBLP:conf/iclr/0001LSXTZPSLSTL25} provide agent-computer interfaces; while Terminal-Bench~\cite{merrill2026terminalbenchbenchmarkingagentshard} and AppWorld~\cite{DBLP:conf/acl/TrivediKHMDLGSB24} evaluate executable-environment interaction. AutoCodeRover~\cite{DBLP:conf/issta/0002RFR24}, Agentless~\cite{DBLP:journals/corr/abs-2407-01489}, and SWE-Dev~\cite{DBLP:journals/corr/abs-2505-16975} target repository repair or feature development. However, these settings typically still define success based on a single task specification and a single trajectory. Our work introduces iterative interaction into this loop. Built on Harbor~\cite{laude2026harbor}, \MYBENCH~ supports persistent environments, cumulative tests, and state accounting across rounds.

\paragraph{Multi-Turn Interaction Evaluation}
\label{sec:rw_multi_turn}

Multi-turn evaluation spans dialogue, general agents, and tool use. MT-Bench~\cite{DBLP:conf/nips/ZhengC00WZL0LXZ23}, MT-Eval~\cite{kwan2024mteval}, and LMSYS-Chat-1M~\cite{zheng2023lmsyschat1m} study conversational interaction; Laban et al.~\cite{laban2026lost} show that distributing information across turns can sharply reduce performance. AgentBench~\cite{DBLP:conf/iclr/0036YZXLL0DMYZ024}, OSWorld~\cite{DBLP:conf/nips/XieZCLZCHCSLLXZ24}, BFCL~\cite{DBLP:conf/icml/PatilMYJSSG25}, $\tau$-bench~\cite{yao2024taubenchbenchmarktoolagentuserinteraction}, and ToolSandbox~\cite{DBLP:conf/naacl/LuHZANBMMLYWP25} evaluate sequential decision making or stateful tool use. In the coding domain, specific studies have investigated multi-turn generation, dependency-ordered code flow, and security~\cite{zheng2025multiturncodegen,wang2025codeflowbench,rawal2025mtsec}. \MYBENCH~ is complementary: it evaluates repository-level development chains by executing cumulative tests after every round, ensuring that each intermediate artifact is behaviorally verified.

\section{\MYBENCH}
\label{sec:Method}

This section defines \MYBENCH~ as a benchmark for persistent multi-turn coding. The main design choice is to treat coding as an evolving workspace rather than isolated prompts: each response changes files, dependencies, interfaces, and tests that later rounds inherit. \MYBENCH~ persists one workspace and one agent session across all rounds, carries active requirements forward through cumulative tests, and records the first round where the accumulated implementation violates the active specification.

\subsection{Task Definition and Taxonomy}
\label{sec:task_def}

\begin{figure*}[t]
\centering
\resizebox{0.92\textwidth}{!}{%
\includegraphics[width=0.85\textwidth]{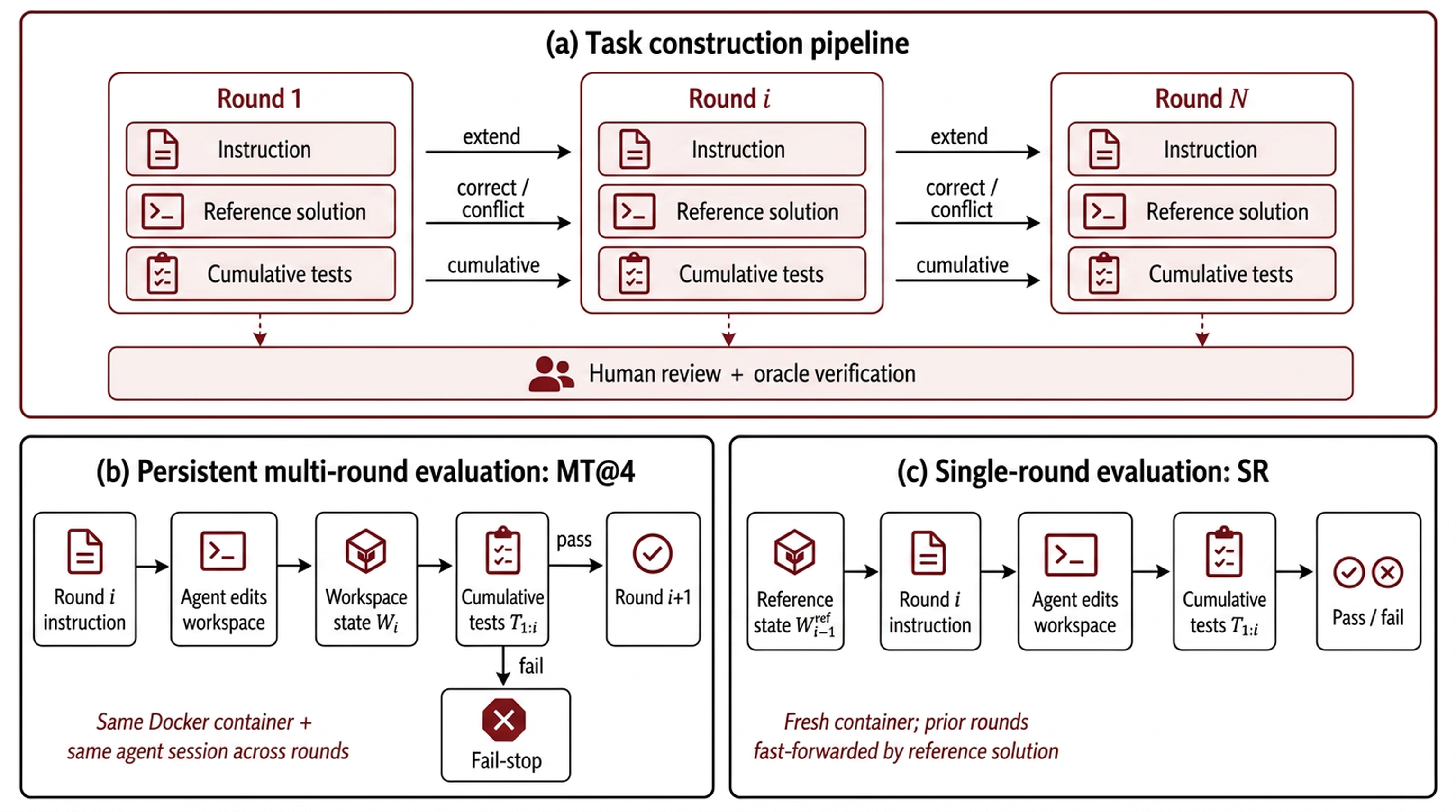}
}
\caption{Overview of \MYBENCH. (a)~Each round contains an instruction, reference solution, and cumulative tests checked by human review and oracle verification. (b)~MT@4 keeps one Docker environment and agent session across rounds, with fail-stop termination. (c)~SR fast-forwards to the reference state before each target round.}
\label{fig:overview}
\end{figure*}

A \MYBENCH~ task has $N$ rounds executed in one persistent Docker container (Figure~\ref{fig:overview}b). At round $i$, the agent receives instruction $\mathcal{I}_i$ and edits workspace $\mathcal{W}_{i-1}$ into $\mathcal{W}_i$. The verifier $\mathcal{T}_{1:i}$ tests all active requirements through round $i$; the next instruction is issued only after a pass. Each round yields $r_i\in\{0,1\}$. If $r_i=0$, evaluation stops and later rounds receive zero credit.

Fail-stop scoring makes the failure point interpretable. Permitting an agent to continue after a failure would conflate two distinct capabilities: recovering from an invalid workspace versus preserving a valid, evolving state. \MYBENCH~ measures the latter in its main multi-turn score; Appendix~\ref{app:scoring_alternatives} discusses recovery-oriented and reference-fast-forwarded alternatives.

\paragraph{Two-dimensional taxonomy.}
Multi-turn coding tasks differ in both communication style and engineering pressure. \MYBENCH~ labels each task along two axes: interaction style and engineering activity.

\paragraph{Interaction styles.}
The first axis captures where requirements live. \textbf{Explorative} tasks start with a detailed request and then use terse follow-ups, testing intent recovery from prior conversation and repository state. \textbf{Contractual} tasks provide detailed behavioral specifications at every round, including revisions to earlier behavior. \textbf{Document-driven} tasks place persistent semantics in project artifacts such as specifications or \texttt{AGENTS.md}, testing whether agents treat repository documents as part of the active specification.

\paragraph{Engineering activities.}
The second axis captures the dominant engineering pressure: construction, specification evolution, review-driven improvement, or migration with compatibility preservation. Table~\ref{tab:cross_product} gives definitions and distribution. Later tables abbreviate these labels as Con, Spec, Rev, Mig, Exp, Ctr, and Doc.

Reporting both axes supports task-level multi-turn analysis and round-level single-round analysis; Appendix~\ref{app:task_examples} gives a released example.

\begin{table*}[t]
\centering
\caption{Task taxonomy and distribution. Each count cell reports \emph{tasks / rounds}, so the table reflects both the multi-round task inventory and the round-level evaluation units used for single-round analysis. Expl., Contr., and Doc. denote explorative, contractual, and document-driven interaction styles.}
\label{tab:cross_product}
\small
\resizebox{0.99\textwidth}{!}{%
\begin{tabular}{p{3.0cm}p{7.0cm}cccc}
\toprule
\textbf{Engineering Activity} & \textbf{Capability measured} & \textbf{Expl.} & \textbf{Contr.} & \textbf{Doc.} & \textbf{Total} \\
\midrule
Construction & Building a system incrementally while preserving earlier features and interfaces. & 9 / 80 & 3 / 37 & 1 / 7 & 13 / 124 \\
Spec Evolution & Updating an implementation after a later round overturns a core assumption. & 1 / 8 & 1 / 7 & 1 / 7 & 3 / 22 \\
Review & Improving non-functional properties such as performance, security, and observability without regression. & 3 / 21 & 1 / 7 & 1 / 9 & 5 / 37 \\
Migration & Moving a legacy system to a new implementation style while keeping backward compatibility. & 3 / 29 & 1 / 7 & 1 / 8 & 5 / 44 \\
\midrule
\textbf{Total} & & \textbf{16 / 138} & \textbf{6 / 58} & \textbf{4 / 31} & \textbf{26 / 227} \\
\bottomrule
\end{tabular}%
}
\end{table*}

\subsection{Data Collection Pipeline}
\label{sec:data_pipeline}

Constructing multi-turn tasks requires coherent requirement chains, cumulative behavioral tests, and incremental reference solutions (Figure~\ref{fig:overview}a). A valid task must make later rounds depend on earlier state while still allowing multiple correct implementation paths. We use a four-stage quality pipeline; Appendices~\ref{app:task_structure} and~\ref{app:qa_details} give the released format and additional QA details.

\paragraph{Task design protocol.}
The main validity risk is that agents may take different early implementation paths. Instructions describe observable behavior, tests verify execution rather than source layout, and each round's tests are independent of prior reference patches. Each task begins with a substantial system foundation and includes at least one correction and one conflict, so agents must both extend and revise existing behavior.

\paragraph{Task construction and internal review.}
Engineers author complete chains, including instructions, reference solutions, tests, and Docker environments. LLM assistance is used for drafting candidate task ideas and failure-analysis prompts, but released instructions, tests, reference deltas, and accepted diagnostic labels are author-reviewed. Two independent reviewers then check instruction coherence and three-way alignment among instructions, tests, and solutions. Reviewers ask whether a competent implementation that differs from the reference can pass, whether each test assertion follows from a current requirement, and whether later rounds preserve all requirements that have not been explicitly superseded. Tasks with critical findings are revised before oracle verification.

\paragraph{Oracle verification.}
Each reference solution is executed in the full environment and must obtain $r_i=1$ for every round. This catches incorrect solutions, missing dependencies, nondeterministic tests, and unsatisfiable specifications.

\paragraph{Cross-validation.}
Final cross-validation verifies reference behavior, traces every assertion to a current specification, and audits shortcut strategies. We inspect whether a model could pass by hard-coding fixture outputs, ignoring documented formats, or satisfying only the latest round while breaking earlier requirements. Unresolved tasks are revised or removed. The resulting benchmark is intentionally smaller than broad single-turn collections, but each released task supplies a longer and more controlled development trajectory.

\subsection{Dataset Statistics}
\label{sec:dataset_stats}

\begin{figure*}[t]
\centering
\resizebox{0.94\textwidth}{!}{%
\begin{tikzpicture}[font=\scriptsize]
  \tikzstyle{panel}=[draw=white,fill=white]

  \begin{scope}[shift={(-0.7,0)}]
    \draw[panel] (0,0) rectangle (4.3,3.3);
    \node[font=\bfseries] at (2.15,3.0) {(a) Task length};
    \foreach \y/\lab/\w/\task/\rounds/\col in {2.15/5--7 rounds/1.50/9/61/PKURed!26,1.45/8--9 rounds/2.00/12/99/PKURed!18,0.75/10--15 rounds/0.85/5/67/PKURose!24} {
      \node[anchor=east] at (0.65,\y+0.13) {\lab};
      \draw[fill=\col,draw=PKUGray!42] (0.85,\y) rectangle ++(\w,0.26);
      \node[anchor=west] at (0.95+\w,\y+0.13) {\task~tasks / \rounds~rounds};
    }
  \end{scope}

  \begin{scope}[shift={(4.8,0)}]
    \draw[panel] (0,0) rectangle (4.55,3.3);
    \node[font=\bfseries] at (2.275,3.0) {(b) Technical domains};
    \foreach \y/\lab/\w/\num/\col in {2.35/ML \& MLOps/2.4/9/PKURed!28,1.95/Data engineering/1.05/4/PKURed!18,1.55/Systems \& code/0.80/3/PKURose!22,1.15/Scientific/0.80/3/PKURose!12,0.75/Testing \& automation/1.05/4/PKUGray!18,0.35/Infra and security/0.80/3/PKUDarkRed!14} {
      \node[anchor=east] at (1.4,\y+0.13) {\lab};
      \draw[fill=\col,draw=PKUGray!42] (1.6,\y) rectangle ++(\w,0.26);
      \node[anchor=west] at (1.65+\w,\y+0.13) {\num};
    }
  \end{scope}

  \begin{scope}[shift={(9.85,0)}]
    \draw[panel] (0,0) rectangle (4.55,3.3);
    \node[font=\bfseries] at (2.275,3.0) {(c) Requirement-change pressure};
    \foreach \y/\lab/\w/\num/\col in {2.05/Extension/3.00/198/PKURed!26,1.35/Correction/1.05/69/PKURose!24,0.65/Conflict/0.65/42/PKUDarkRed!16} {
      \node[anchor=east] at (1.05,\y+0.13) {\lab};
      \draw[fill=\col,draw=PKUGray!42] (1.25,\y) rectangle ++(\w,0.26);
      \node[anchor=west] at (1.35+\w,\y+0.13) {\num};
    }
    \node[align=center] at (2.275,0.18) {110 rounds carry $\geq$1 correction or conflict};
  \end{scope}
\end{tikzpicture}%
}
\caption{Dataset statistics for \MYBENCH~ beyond the taxonomy distribution in Table~\ref{tab:cross_product}. The benchmark contains 26 multi-turn tasks and 227 evaluated rounds, with 5 to 15 rounds per task, broad technical coverage, and frequent requirement corrections or conflicts.}
\label{fig:stats}
\end{figure*}
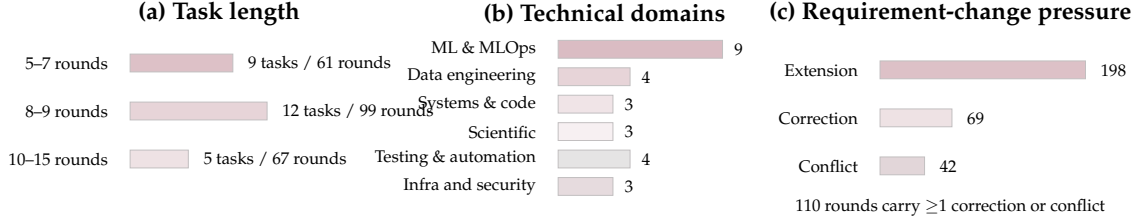

\MYBENCH~ contains 26 tasks and 227 evaluated rounds. Figure~\ref{fig:stats} summarizes scale, task length, technical coverage, and requirement-change pressure; Table~\ref{tab:cross_product} gives the taxonomy cross product. Each task is long enough for early implementation choices to constrain later work, while short enough for failures to be traced to concrete requirement changes.

The benchmark covers MLOps, data engineering, systems programming, scientific computing, testing, automation, cloud/devops, security, and compiler implementation. We intentionally avoid treating domain labels as the primary taxonomy because technical domain and multi-turn behavior are not the same property: an MLOps task can be construction-oriented, review-driven, or migration-oriented depending on the requirement chain. Domain coverage instead supports external validity, while the two-dimensional taxonomy supports analysis of multi-turn behavior.

Rounds are annotated with non-exclusive change types: \emph{extension}, \emph{correction}, and \emph{conflict}. The distinction captures stale behavior after specification changes, not only missing latest-round features.

Instruction length follows interaction style: explorative tasks shorten after the first round, contractual tasks stay detailed, and document-driven tasks move semantics into artifacts such as \texttt{AGENTS.md}. All styles use the same executable verifier.

\subsection{Evaluation Protocol}
\label{sec:eval_protocol}

All evaluations use Harbor with the Terminus-2 agent harness, which gives each model shell access inside a Docker container. In multi-round evaluation, the same container and agent session persist across rounds. Harbor handles round sequencing, cumulative verification, snapshotting, SR reference fast-forwarding, and fail-stop accounting, turning a single-instruction trial into a round-indexed trial with continuous workspace, session, reward log, and resume lineage. We use the same harness, environment, tool interface, and prompt protocol for all models; Appendix~\ref{app:terminus2} gives the exact configuration, and Appendix~\ref{app:harbor_extensions} details the Harbor multi-turn extensions.

For task $t$ with $N_t$ rounds, let $r_{t,a,i}\in\{0,1\}$ be the cumulative verifier reward for attempt $a$ at round $i$. Under fail-stop, if attempt $a$ first fails at round $i$, then $r_{t,a,j}=0$ for every $j>i$. The single-attempt multi-turn score is
\begin{equation}
  S_{t,a} = \frac{1}{N_t}\sum_{i=1}^{N_t} r_{t,a,i}.
\end{equation}
We report \textbf{MT@4}, defined as $\frac{1}{|\mathcal{D}|}\sum_t \frac{1}{N_t}\sum_i \max_{a\le4} r_{t,a,i}$. Equivalently, MT@4 gives credit for a round if any of four attempts reaches that round with a workspace that still satisfies the cumulative verifier. \textbf{SR} is $\frac{1}{\sum_t N_t}\sum_{t,i}s_{t,i}$, where $s_{t,i}$ is the binary reward for solving round $i$ after Harbor fast-forwards earlier rounds with reference deltas. SR measures a different condition from MT@4: whether the agent can solve a target round when all earlier work has been completed by the reference solution. A large MT@4--SR gap means that isolated instruction-following ability is not enough, under this protocol, to maintain the agent's own long-horizon workspace.

\textbf{Comp} is $\frac{1}{|\mathcal{D}|}\sum_t \mathbf{1}[\max_{a\le4} r_{t,a,N_t}=1]$, the fraction of tasks completed through the final round in at least one attempt. We also report \textbf{Avg.\ Turns} and \textbf{Output Tok.\ (K)} for multi-round runs, with partial trajectories scaled to the full task horizon. Each $\mathcal{T}_{1:i}$ tests all still-valid requirements through round $i$ and detects regressions. Appendix~\ref{app:infrastructure} gives exact metric accounting.

\section{Experiments}
\label{sec:Evaluation}

\subsection{Evaluation Setup}
\label{sec:setup}

We evaluate the 13 coding agents in Table~\ref{tab:main_results} under a shared protocol. Each agent-task pair runs in a dedicated Docker container, and correctness is measured only by executing the task verifier. Multi-round evaluation keeps the same workspace and agent session across rounds and uses four independent attempts. Single-round evaluation fast-forwards the workspace to the reference state before the target round. Appendix~\ref{app:model_config} maps compact model labels to full names and endpoint identifiers.

\MYBENCH~ contains 26 tasks and 227 evaluated rounds, with 5 to 15 rounds per task. We aggregate by agent, engineering activity, interaction style, and round index to distinguish single-round performance, persistent execution, and degradation over longer requirement chains.

We report \textbf{MT@4}, \textbf{SR}, \textbf{Comp}, \textbf{Avg.\ Turns}, and \textbf{Output Tok.\ (K)} as defined in Section~\ref{sec:eval_protocol}, and sort the main table by MT@4. MT@4 asks whether any of four attempts can maintain a valid evolving workspace; SR asks whether a single attempt can solve the same round from a reference-completed state. Together they separate accumulated-state reliability from isolated instruction-following ability.

\subsection{Main Results}
\label{sec:main_results}

\begin{table*}[t]
\centering
\caption{Main results on \MYBENCH. Rows are sorted by \textbf{MT@4}. \textbf{MT@4} is the four-attempt fail-stop multi-round score, \textbf{SR} is reference-fast-forward single-round pass rate, and \textbf{Comp} is full-task completion. \textbf{Avg. Turns} is agent--model exchanges per full task; \textbf{Output Tok. (K)} is generated-token usage in thousands. Activity columns abbreviate construction, specification evolution, review-driven improvement, and migration; style columns abbreviate explorative, contractual, and document-driven tasks.}
\label{tab:main_results}
\resizebox{0.975\textwidth}{!}{%
\begin{tabular}{l ccccc cccc ccc}
\toprule
 & \multicolumn{5}{c}{\textbf{Overall}} & \multicolumn{4}{c}{\textbf{Engineering Activity}} & \multicolumn{3}{c}{\textbf{Interaction Style}} \\
\cmidrule(lr){2-6} \cmidrule(lr){7-10} \cmidrule(lr){11-13}
\rowcolor{PKULightRed}
\textbf{Agent} & \textbf{MT@4} & \textbf{SR} & \textbf{Comp} & \makecell{\textbf{Avg.}\\\textbf{Turns}} & \makecell{\textbf{Output Tok.}\\\textbf{(K)}} & \textbf{Con} & \textbf{Spec} & \textbf{Rev} & \textbf{Mig} & \textbf{Exp} & \textbf{Ctr} & \textbf{Doc} \\
\midrule
Opus-4.7 & \textbf{54.0} & 76.7 & \textbf{42.3} & 590.6 & 50.0 & \textbf{58.8} & 42.9 & \textbf{70.0} & 32.3 & 41.8 & \textbf{64.1} & 82.1 \\
GPT-5.5 & 52.4 & 74.4 & 38.5 & 456.3 & 74.1 & 57.3 & 33.3 & 50.0 & \textbf{53.6} & \textbf{51.0} & 42.6 & 75.0 \\
Opus-4.6 & 44.0 & \textbf{78.9} & 34.6 & 747.5 & 734.2 & 37.4 & \textbf{66.7} & 44.0 & 47.9 & 33.1 & 35.5 & \textbf{100.0} \\
GLM-5.1 & 36.2 & 63.9 & 15.4 & 859.8 & 104.2 & 29.6 & 61.9 & 43.1 & 30.9 & 23.3 & 30.5 & 94.4 \\
Kimi-K2.6 & 31.9 & 59.0 & 23.1 & 1155.5 & 92.5 & 29.5 & 33.3 & 42.5 & 26.9 & 21.6 & 29.5 & 75.0 \\
DS-V4-Pro & 30.6 & 56.4 & 19.2 & 1134.8 & 168.8 & 37.3 & 0.0 & 33.0 & 29.4 & 23.7 & 20.1 & 75.0 \\
Qwen3.6-Plus & 29.4 & 57.3 & 15.4 & 629.3 & 103.1 & 30.2 & 45.8 & 22.5 & 24.3 & 28.6 & 19.3 & 50.0 \\
MiMo-V2.5 & 17.3 & 7.9 & 11.5 & 754.8 & 125.7 & 20.6 & 0.0 & 22.5 & 13.6 & 17.6 & 12.1 & 25.0 \\
Gemini-3.1 & 13.7 & 46.7 & 11.5 & 261.3 & 72.7 & 12.0 & 0.0 & 20.0 & 20.0 & 10.4 & 0.0 & 50.0 \\
DS-V4-Flash & 9.4 & 46.3 & 0.0 & 1104.7 & 148.7 & 5.2 & 0.0 & 28.6 & 6.9 & 9.4 & 7.1 & 13.9 \\
Qwen3.5-397B & 4.6 & 44.1 & 0.0 & 587.8 & 53.0 & 5.4 & 0.0 & 2.2 & 7.9 & 4.4 & 6.1 & 2.8 \\
MiniMax-M2.7 & 3.7 & 30.0 & 0.0 & 600.4 & 59.2 & 2.2 & 0.0 & 6.7 & 6.9 & 3.3 & 2.0 & 8.3 \\
Doubao-2.0 & 1.9 & 23.8 & 0.0 & 211.1 & 18.5 & 2.9 & 0.0 & 0.0 & 2.5 & 3.3 & 0.0 & 0.0 \\
\bottomrule
\end{tabular}%
}
\end{table*}

\begin{figure*}[t]
\centering
\includegraphics[width=0.92\textwidth]{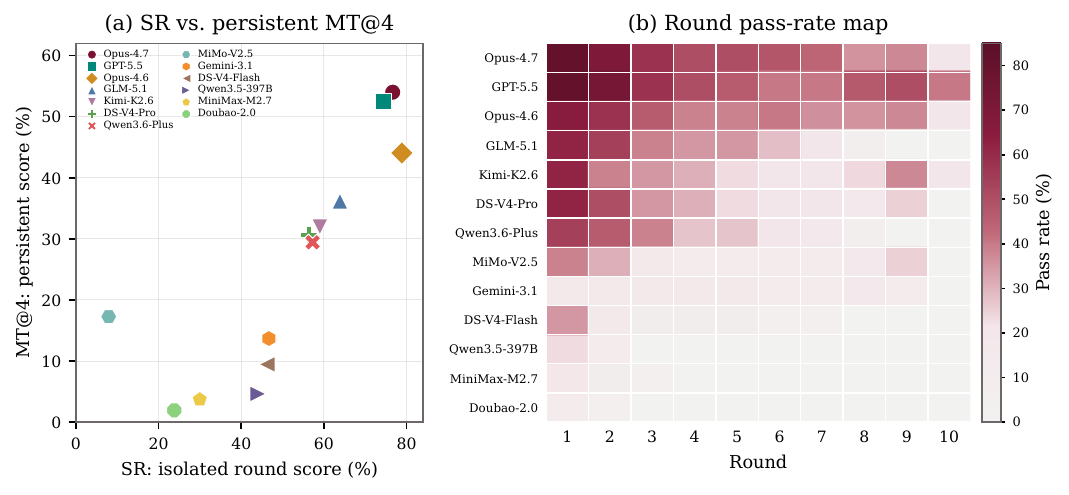}
\caption{(a)~SR vs.\ persistent MT@4 for each agent. (b)~Per-round pass rate heatmap; rows are agents sorted by MT@4, columns are rounds.}
\label{fig:main_results}
\end{figure*}

\paragraph{The benchmark is challenging even for the strongest agents.}
Opus-4.7 leads with 54.0 MT@4 and GPT-5.5 follows at 52.4; only these two agents exceed half of the total multi-round credit. Opus-4.6 reaches 44.0, and every other agent scores at or below 36.2. Full-task completion rates are correspondingly low: Opus-4.7 completes 42.3\% of tasks through the final round, GPT-5.5 38.5\%, and no agent below the top three exceeds 23.1\%.

\paragraph{Multi-turn execution widens the gap across tiers.}
Grouping agents into top, middle, and lower tiers by MT@4, the top-to-lower ratio is 5.9$\times$ (50.1 vs.\ 8.4), compared with only 2.3$\times$ under SR (76.7 vs.\ 33.1). Persistent execution separates agents more sharply than single-round evaluation. All agents except MiMo-V2.5-Pro have SR above MT@4 by 21.9--39.5 points; Opus-4.6 is the clearest reranking, with the highest SR (78.9) but only the third-highest MT@4 (44.0).

The gap is not an artifact of giving MT@4 four attempts and SR one attempt. At round~1, the best-of-four advantage makes MT@4 exceed SR; from round~2 onward, the ordering reverses (Appendix~\ref{app:sr_mt_round}). After the first round, accumulated state outweighs the retry advantage. MiMo-V2.5-Pro is the sole exception (SR 7.9 vs.\ MT@4 17.3), largely because 60\% of its multi-turn failures occur at round~1. This case illustrates that SR rewards extending a reference-completed codebase, whereas MT@4 rewards building and maintaining one's own.

\begin{figure*}[t]
\centering
\includegraphics[width=0.88\textwidth]{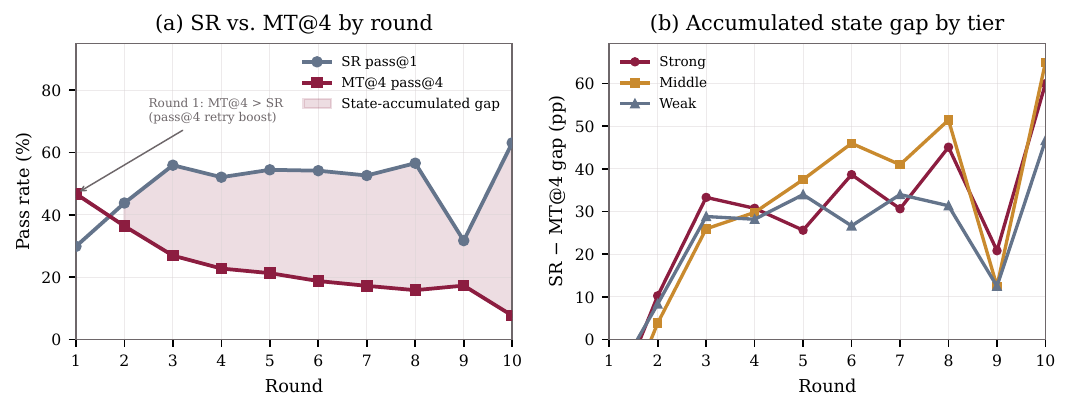}
\caption{Per-round SR (single attempt) vs.\ MT@4 (best of four attempts). (a)~SR remains stable while MT@4 declines; the shaded region is the state-accumulated gap. (b)~SR--MT@4 gap by agent tier; late-round fluctuations reflect small task counts.}
\label{fig:sr_vs_mt_round}
\end{figure*}

\paragraph{Performance degrades sharply across rounds.}
Round-level MT@4 pass rates fall from 46.7 at round~1 to 26.9 at round~3, 21.3 at round~5, and 7.7 at round~10. Some decline reflects the smaller set of long tasks, but the within-task heatmap in Figure~\ref{fig:main_results}(b) also shows falling pass rates for most agents. SR further separates round difficulty from persistent-state degradation: it remains stable at 52--57\% for rounds~3--8 while MT@4 continues to decline (Figure~\ref{fig:sr_vs_mt_round}). The SR--MT@4 gap reaches 33 points by round~5 and 41 points by round~8, indicating that accumulated workspace state drives much of the multi-turn drop. A controlled comparison reinforces this: 57.0\% of failed MT@4 round records are solvable under SR from a reference-completed state (Appendix~\ref{app:workspace_penalty}). The penalty grows with depth: only 15.0\% of round-1 MT failures are SR-solvable, rising to 59.0\% at round~7 and above 80\% beyond round~12 (Figure~\ref{fig:workspace_state_penalty}a).

\paragraph{Cross-attempt variance increases with round depth.}
MT@4 credits a round if any of four attempts passes, but this masks cross-attempt variability. Decomposing into \emph{aptitude} (any attempt passes) and \emph{full consistency} (all four pass), the reliability ratio drops from 67\% at round~1 to 20\% at round~5 (Appendix~\ref{app:variance}), echoing~\cite{bsharat2025llms}: multi-turn settings degrade consistency more sharply than capability. The location of first failure also differs by tier: 57.4\% of lower-tier trial failures occur in the first 20\% of rounds, compared with 29.4\% for mid-tier and 14.4\% for top-tier agents (Appendix~\ref{app:first_fail}), consistent with the view that lower-tier agents lack basic requirement satisfaction while higher-tier agents survive longer before encountering destabilizing corrections or conflicts.

\subsection{Fine-Grained Analysis}
\label{sec:analysis}

\paragraph{Failure patterns are tier-dependent.}
We label each executed failed round by its primary failure mode: missed active requirement, environment/tooling failure, regression, stale superseded behavior, or context loss. Missed-requirement dominates every tier (87--90\%), so the more informative diagnostic is \emph{which secondary modes emerge and when}. Mid-tier failures include stale behavior (28.6\%) from superseded requirements left in the codebase; top-tier failures at advanced rounds include regression (18.2\%) where previously passing tests break after new edits. Per-round profiles (Appendix~\ref{app:failure_progression}) show that for top-tier agents regression peaks at round~2 (35\%) and for mid-tier agents conflict-mishandling appears at round~5 (23\%), while lower-tier agents are dominated by missed requirements throughout.

These tier differences are partly a selection effect: lower-tier agents fail before encountering the correction and conflict rounds where stale-behavior and regression failures become possible, so their failure distribution is truncated rather than qualitatively different. Mid-tier agents survive enough rounds to encounter specification supersession but fail to update already-implemented behavior accordingly, suggesting that specification tracking across accumulating requirements is a larger bottleneck than basic coding ability at this tier. Top-tier agents survive the longest but show regression when editing code to satisfy new requirements. This failure mode appears only after the agent has built enough working functionality for later edits to break.

Appendix~\ref{app:failure_annotation} defines the taxonomy; Appendix~\ref{app:case_study} presents case studies.

\paragraph{Taxonomy-level patterns.}
Across interaction styles, document-driven tasks average 50.1 MT@4, roughly 2.4$\times$ the mean of explorative (20.9) and contractual (20.7) tasks. Across engineering activities, review-driven improvement scores highest (29.6) while specification evolution scores lowest (21.8); leaders also vary by activity, with Opus-4.7 leading construction and review, GPT-5.5 migration, and Opus-4.6 specification evolution. These patterns are diagnostic rather than stable rankings because small cells and task content co-vary with taxonomy labels. When change-type effects are controlled for round position (Appendix~\ref{app:change_type_controlled}), extension rounds show notably higher MT@4 pass rates than correction or conflict rounds in early rounds (43.2\% vs.\ 31.9\% vs.\ 25.0\%), with the difference diminishing in later buckets, indicating that requirement revision adds difficulty beyond later round position.

\paragraph{Failing rounds are associated with higher token consumption.}
To control for the confound that later rounds have longer contexts, we compare output tokens of passing and failing trials at the same round index. Across rounds~1--9, where both pass and fail samples are sufficient, failing trials produce 1.1--3.1$\times$ as many output tokens as passing trials at the same round (Appendix~\ref{app:token_diagnostics}). The direction of this association is ambiguous: harder workspace states may cause both failure and increased agent effort. We report this pattern as a diagnostic observation, not a causal claim.

\section{Discussion}
\label{sec:Discussion}

Section~\ref{sec:Evaluation} shows that persistent multi-turn coding is not only longer single-round coding. The same instructions are often solvable from a reference-completed state but fail when the agent relies on its own accumulated workspace. This distinction matters as coding agents are deployed as ongoing development partners rather than one-shot patch generators.

\paragraph{Implications for agent system designers.}
The SR--MT@4 gap implies that reliability cannot be obtained by improving latest-request execution alone. Agents must preserve a running contract over files, tests, dependencies, and earlier corrections. Appendix~\ref{app:workspace_penalty} shows that 57.0\% of multi-turn failures involve rounds solvable from a reference-completed state, rising above 80\% at deeper rounds and peaking for correction (73.5\%) and conflict (67.1\%) rounds. This suggests useful system support around the model: regression checks between rounds, repository-grounded summaries of active requirements, and workspace audits that detect stale or superseded behavior.

\paragraph{Implications for benchmark designers.}
\MYBENCH~also suggests benchmark-design principles that may transfer beyond coding. Cumulative verification is needed because round-local tests miss regressions against still-active requirements. Fail-stop MT@4 and reference-fast-forward SR should be read together: one measures persistent reliability, while the other estimates whether the target round is solvable absent accumulated workspace damage. Behavior-only tests are also important because correct agents may take different implementation paths across attempts.

Fail-stop scoring is strict, but it gives a clean interpretation to the failure point: the workspace has stopped satisfying the cumulative active specification.
This is useful for measuring persistence, because later instructions are difficult to interpret once the artifact is already invalid.
At the same time, fail-stop scoring deliberately does not measure recovery.
An agent might be able to repair a broken workspace after receiving diagnostics, ask clarification questions, or restart from a clean state, but those are different capabilities.
Benchmarks that care about recovery should add a separate recovery protocol rather than mixing repair into the primary persistence score.
Otherwise later-round failures become ambiguous: they may reflect poor maintenance, weak repair, or difficulty with the new instruction.

\paragraph{Implications for model developers.}
The tier-dependent failures suggest different training bottlenecks. Lower-tier agents need stronger basic requirement satisfaction; mid-tier agents need better specification tracking and conflict resolution; top-tier agents need better regression control during incremental edits. The results argue against treating multi-turn coding as only a context-length problem: longer context helps only if the model can identify active requirements, superseded requirements, and earlier implementation choices that have become liabilities. Training on corrections, conflicts, migrations, and cumulative regression checks exercises behaviors that single-turn coding data often hides.

\paragraph{Implications for practitioners and end users.}
For practitioners, the SR--MT@4 gap suggests that clean workspace restarts or reference-state repair may recover performance, though \MYBENCH~does not directly evaluate restart strategies. The evaluation consumed approximately 24.4 billion tokens across 3{,}657 logged multi-round execution records (Appendix~\ref{app:token_diagnostics}), which makes persistent multi-turn evaluation expensive. MT@4 and SR should be read together: high SR with much lower MT@4 indicates strength on isolated requests but degradation under sustained iterative development.

\section{Conclusion}
\label{sec:Conclusion}

We introduced \MYBENCH, a benchmark for persistent multi-turn coding with cumulative tests, stateful workspaces, and round-level diagnostics. The results show that single-round and multi-turn evaluations measure different conditions: the former rewards extending a reference-completed codebase, while the latter rewards building and maintaining one's own. SR exceeds MT@4 by 22 to 40 points for most agents, and the highest single-round scorer ranks only third in persistent execution. Only two agents exceed 50 MT@4. The failure patterns also differ by tier: lower-tier agents miss early requirements, while top-tier agents more often fail on corrections and regressions.

\bibliographystyle{colm2024_conference}
\bibliography{reference}

\newpage

\appendix

\section{Limitations and Future Work}
\label{app:limitations}
The benchmark contains 26 tasks; scaling is mainly constrained by the cost of manual instruction authoring, cumulative test construction, cross-validation, and multi-agent evaluation. The fail-stop model may underestimate agents capable of recovery, though the SR metric partially addresses this. \MYBENCH~ covers only tasks for which correctness can be verified through executable tests; evaluating agent reliability on subjective design decisions, code quality judgments, and collaborative negotiations remains an open problem.

A natural next step is adaptive multi-turn evaluation: an examiner agent maintains a target system specification and generates instructions from the evaluated agent's current workspace state and performance trajectory. This would remove the fixed round structure and let the evaluation adapt its difficulty and direction to expose each agent's failure boundary. It may also reduce the manual cost of task construction. Extending this approach beyond coding to other tool-using domains, where each turn modifies a shared artifact, remains open.

\section{Task Format and Reproducibility Details}
\label{app:task_structure}

Section~\ref{sec:Method} defines the benchmark semantics: multi-round task execution, cumulative verification, fail-stop scoring, and the two-dimensional taxonomy. This appendix provides the exact release format and execution details needed for reproducibility. All tasks follow the same directory layout, metadata schema, and execution protocol described below.

\subsection{Released Task Layout}

Each task is a self-contained directory with one environment definition and $N$ round subdirectories:

\begin{verbatim}
task/
  task.toml                    # Metadata
  environment/
    Dockerfile                 # Shared runtime environment
  round_1/
    instruction.md             # Round 1 instructions
    solution/solve.sh          # Round 1 reference solution
    tests/test.sh              # Round 1 verification (cumulative)
  round_2/
    instruction.md
    solution/solve.sh
    tests/test.sh
  ...
  round_N/
    instruction.md
    solution/solve.sh
    tests/test.sh
\end{verbatim}

The \texttt{Dockerfile} in the \texttt{environment/} directory defines a reproducible runtime that includes all language runtimes, system libraries, and tool dependencies needed by the task. Each round's \texttt{instruction.md} is the verbatim instruction delivered to the agent. The \texttt{solve.sh} script applies the reference solution for that round only, and the \texttt{test.sh} script runs the cumulative verifier.

\subsection{Task Metadata Format}

The \texttt{task.toml} file records all metadata needed to configure evaluation:

\begin{verbatim}
[metadata]
name = "protocol-parser-evolution"
num_rounds = 5
engineering_activity = "spec_evolution"
interaction_style = "document_driven"
change_types = [
  ["extension"],
  ["extension", "correction"],
  ["conflict"],
  ["conflict", "extension"],
  ["extension"]
]
technology_domain = "system_protocol"
has_agents_md = false
\end{verbatim}

The \texttt{change\_types} array records the requirement-change labels for each round. Valid labels are \textbf{extension} (adding new capabilities without altering existing behavior), \textbf{correction} (adjusting previously established behavior without overturning core assumptions), and \textbf{conflict} (overturning a core assumption and requiring structural adaptation). A single round may carry multiple labels when the instruction combines, for example, an extension with a correction. These labels are used for the diagnostic analyses in Section~\ref{sec:analysis} and for the change-type breakdowns in Appendix~\ref{app:full_results}.

\subsection{Execution Semantics}

All rounds execute within a single Docker container. The evaluation framework builds the Docker image from the task's \texttt{Dockerfile}, initializes one agent session for the entire task, and then executes each round sequentially: delivering the instruction, waiting for the agent to signal completion, running the cumulative test script, and recording the binary reward. If any round fails ($r_i = 0$), the task terminates immediately under the fail-stop protocol described in Section~\ref{sec:task_def}. After all rounds complete, the composite score is computed as $S = \frac{1}{N}\sum_{i=1}^{N} r_i$. This execution model ensures that every evaluated round depends on the workspace state produced by the same agent in all preceding rounds, which is the defining property of persistent multi-turn evaluation.

\subsection{Reference Solution Semantics}

Each round's \texttt{solve.sh} makes only that round's incremental changes. Executing \texttt{solve.sh} for rounds 1 through $N$ in sequence produces the complete reference state after round $N$. Reference solutions must never rewrite the entire project; they build incrementally on the prior state, mirroring the incremental development process that the benchmark requires of evaluated agents. This constraint is verified during oracle verification (Section~\ref{sec:data_pipeline}): each reference delta is applied in sequence, and the cumulative verifier must pass at every round boundary.

\subsection{Cumulative Test Semantics}

Round $i$'s \texttt{test.sh} verifies all still-valid requirements from rounds 1 through $i$. When a later round's conflict or correction supersedes an earlier requirement, the cumulative test is updated accordingly: assertions that check the now-obsolete behavior are replaced by assertions that check the new behavior. Tests run all assertions to completion without early termination on the first failure, producing a pass/fail report that supports diagnostic failure annotation (Appendix~\ref{app:failure_annotation}). Test scripts verify behavior through actual system execution only: they invoke the built system, send inputs, check outputs, and compare results. They never inspect source code structure, function names, file organization, or class hierarchies, so the verifier can accept behaviorally correct implementations with different internal designs.

\section{Task Examples}
\label{app:task_examples}

Section~\ref{sec:dataset_stats} reports that \MYBENCH~ contains 26 tasks spanning 227 evaluated rounds. This appendix presents one released task in abbreviated form to illustrate the cumulative verification design and the cross-round dependencies that distinguish multi-turn evaluation from concatenated single-turn tasks.

\subsection{Released Task: Deterministic Data Pipeline CLI}
\label{app:example_dpipe}

The task \texttt{deterministic-data-pipeline-go} is a 15-round contractual construction task in the data-engineering-reproducibility domain. The agent must build and extend \texttt{dpipe}, a deterministic Go CLI for CSV ingestion, transformation, validation, profiling, lineage tracking, quality checks, snapshots, and changelogs. It is representative because it combines cumulative feature growth, behavior corrections, and explicit conflicts with earlier output formats. Table~\ref{tab:dpipe_example} shows an abbreviated round chain.

\begin{table}[htbp]
\caption{Abbreviated round chain from the released task \texttt{deterministic-data-pipeline-go}. The omitted rounds add commands such as profiling, validation, audit logging, schema inspection, fingerprinting, snapshots, and changelogs.}
\label{tab:dpipe_example}
\centering
\small
\begin{tabular}{cL{2.0cm}L{9.7cm}}
\toprule
\textbf{Round} & \textbf{Change type} & \textbf{Instruction summary} \\
\midrule
1 & Extension & Build \texttt{dpipe} as a deterministic Go CLI that reads CSV files, validates rows against JSON schemas, applies ordered transformations, writes bit-identical outputs, and records reproducibility metadata. \\
3 & Extension, correction & Correct linear-fill boundary behavior and z-score normalization while adding extended profiling and a new \texttt{drift} command. The verifier checks both the new behavior and the earlier deterministic pipeline. \\
5 & Extension, conflict & Change the \texttt{verify} and \texttt{manifest} checksum behavior from SHA-256 sidecar files to BLAKE2b-256 manifest checksums, while adding a \texttt{lineage} command. \\
7 & Extension, correction & Correct lineage semantics for manifest steps: directory scans should not become input nodes. Add a temporal transform for time-series operations. \\
10 & Extension, conflict & Replace the manifest entry field \texttt{blake2b} with \texttt{checksum} and \texttt{algorithm}, then add a \texttt{reconcile} command that compares two manifests. \\
12 & Extension, correction & Correct the default \texttt{normalize} method so an omitted or empty method behaves as \texttt{minmax}. Add a numeric \texttt{round} transform. \\
15 & Extension, conflict & Change \texttt{snapshot} row hashes from MD5 to SHA-256 and add a constant-value \texttt{tag} transform. The final verifier exercises the accumulated system across transforms, auditability, lineage, manifest formats, snapshots, and changelogs. \\
\bottomrule
\end{tabular}
\end{table}

This example illustrates why the benchmark uses cumulative verification. A correct round 15 submission must do more than implement the latest hash-format change. It must also preserve deterministic output, earlier checksum semantics where still active, profile/validate/quality/schema behavior, audit logging, lineage metadata, and snapshot/changelog requirements introduced across the previous rounds. This is the evaluation setting targeted by \MYBENCH: later instructions may look local, but the verifier scores the whole evolving system.

\subsection{Document-Driven Constraints}
\label{app:agentsmd_example}

Section~\ref{sec:task_def} introduces document-driven interaction as one of the three interaction styles. In document-driven tasks, persistent requirements are encoded in project artifacts such as \texttt{AGENTS.md} or specification files rather than in the chat-based instruction stream. The instruction for a given round may simply state that the specification file has been updated, without repeating the changed requirements in the instruction text. The cumulative verifier then checks whether the agent's implementation is synchronized to the updated artifact.

This design tests a capability that purely chat-centric evaluation cannot measure: whether the agent reads, interprets, and preserves repository-level constraints across rounds. In practice, the benchmark's four document-driven tasks each contain an evolving specification document. A round may add a new constraint, tighten an existing tolerance, or remove a previously required behavior through a specification update. The verifier checks the implementation against the current state of the specification, not against the instruction text alone. Agents that rely exclusively on the latest chat message and ignore changes to project files are expected to fail on these tasks, particularly after specification updates that are not echoed in the instruction.

\section{Quality Assurance Details}
\label{app:qa_details}

Section~\ref{sec:data_pipeline} describes the four-stage quality pipeline used to construct \MYBENCH~ tasks. This appendix provides additional detail on the design principles, cross-validation protocol, and failure annotation methodology.

\subsection{Design Principles}

Three principles govern all task construction in \MYBENCH. They address a basic challenge of multi-turn benchmark design: different agents may reach equivalent behavior through different implementation paths, so neither instructions nor tests can assume a specific internal design.

\begin{enumerate}[leftmargin=*, nosep]
  \item \textbf{Describe behavior, not implementation.} Instructions specify what the system should do in terms of observable behavior, not how to achieve it. No algorithm names, data structure choices, or file organization prescriptions appear in instructions. This lets different agents reach equivalent behavioral goals through different implementation paths.
  \item \textbf{Test behavioral requirements, not implementation details.} Verification scripts check the system's external behavior through actual execution: running commands, sending inputs, and checking outputs. They never inspect source code structure, function names, or file organization. This makes the same test suite valid across arbitrary implementation paths.
  \item \textbf{Round-independent testing.} Round $i$'s tests cannot assume the agent used the same code structure as the reference solution in rounds 1 through $i-1$. They can only assume the agent passed the behavioral tests of all prior rounds. This handles divergent implementation paths in multi-turn evaluation.
\end{enumerate}

These principles are enforced at every stage of the pipeline: during task authoring, during internal review, during oracle verification, and during final cross-validation. Violations discovered at any stage trigger task revision before the task enters the benchmark.

\subsection{Cross-Validation Protocol}

After tasks pass internal review and oracle verification, a final cross-validation stage addresses three objectives that go beyond checking whether the reference solution is correct:

\begin{enumerate}[leftmargin=*, nosep]
  \item \textbf{Answer correctness verification.} Independent reviewers verify that the reference solution for each round produces the behavior specified by the instructions. They check for alternative valid interpretations that the tests might not distinguish and for edge cases that the cumulative tests might miss.
  \item \textbf{Test-specification alignment audit.} Reviewers verify that every test assertion traces back to a current specification, either newly introduced in the current round or persisting from a previous round. They also verify that no test checks stale requirements that have been superseded by later rounds, which would cause otherwise correct implementations to fail.
  \item \textbf{Shortcut analysis.} Reviewers attempt to identify implementation strategies that would pass all tests without satisfying the task requirements. This includes checking for overly specific test inputs that allow hardcoding, test orderings that leak information, and behavioral checks that are too loose to distinguish correct from incorrect implementations. Tasks with identified shortcuts are revised to close the gap.
\end{enumerate}

Tasks that cannot be brought into alignment across all three objectives are removed from the benchmark. This conservative filtering contributes to the benchmark's relatively small size (26 tasks) but ensures that each released task provides a controlled evaluation trajectory.

\subsection{Failure Annotation Protocol}
\label{app:failure_annotation}

Section~\ref{sec:analysis} reports tier-dependent failure patterns. To produce these diagnostics, we build one evidence packet for each executed multi-round fragment whose verifier reward is zero. We do not classify implicit zeros from unexecuted future rounds (which result from fail-stop termination). Each packet contains the model identity, task taxonomy labels, failed round index, change types, current and previous instructions, verifier stdout/stderr, test exit code, cumulative test script excerpts, reference-solution excerpts, and paths back to the raw Harbor result. Verifier output is placed first in the annotation prompt so that concrete assertion failures remain visible under context-length truncation.

An LLM annotator assigns one primary label, optional secondary labels, a confidence score, and short supporting evidence. The five reported labels are:
\begin{itemize}[leftmargin=*, nosep]
  \item \emph{Missed active requirement}: the failed assertions correspond to behavior newly required or still required at the failed round, and the submitted workspace does not implement that behavior.
  \item \emph{Environment/tooling failure}: package installation, build setup, command wiring, generated artifacts, permissions, or runtime invocation prevents the verifier from exercising the intended behavior.
  \item \emph{Regression}: behavior that had become active in an earlier round is broken by a later edit, even though the latest instruction may have been partially implemented.
  \item \emph{Stale superseded behavior}: a conflict round explicitly replaces earlier behavior, but the workspace continues to expose the obsolete behavior.
  \item \emph{Context loss}: the evidence shows that the agent ignored necessary prior task state, files, or corrected semantics; this label is used only when the failure is not better explained as a local missed requirement.
\end{itemize}
The prompt requires \texttt{unknown} when logs do not identify a root cause. Low-confidence, malformed, timeout, and \texttt{unknown} annotations are retained in a review queue and excluded from aggregate failure-mode percentages. The accepted labels are used as evidence-backed qualitative diagnostics, not as an additional scoring metric.

\paragraph{Human validation.}
All failure annotations from the multi-turn evaluation were manually reviewed by the authors; every accepted label was confirmed against the verifier output and agent trajectory. For the single-round evaluation, we sampled 50 annotated failures uniformly across tiers and failure labels and independently re-annotated them. The human labels agreed with the LLM annotations on all 50 cases (100\% agreement), indicating that the structured evidence packets provide sufficient information for reliable classification.

\begin{figure*}[t]
\centering
\includegraphics[width=0.86\textwidth]{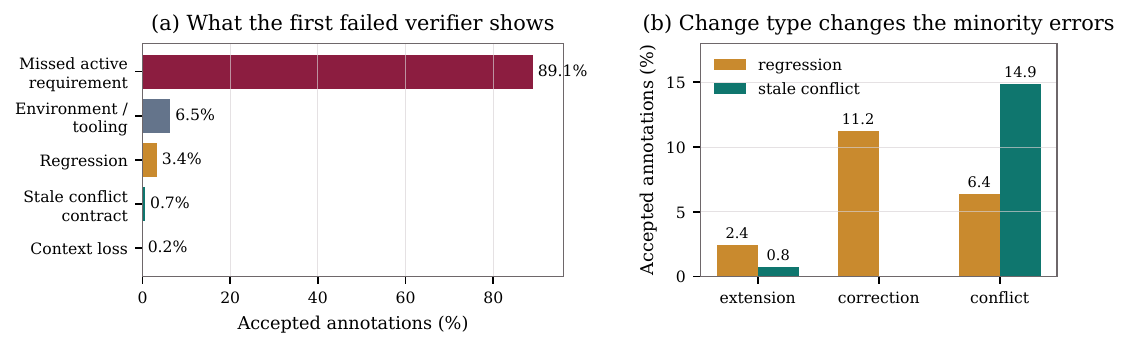}
\caption{Failure diagnostics expanded from Section~\ref{sec:analysis}. (a)~The primary accepted label for each executed failed round fragment. (b)~Regression and stale-conflict labels by task-metadata change type, showing why correction and conflict rounds expose different failure surfaces.}
\label{fig:appendix_failure_story}
\end{figure*}

Figure~\ref{fig:appendix_failure_story} expands the tier-level failure diagnostics from Section~\ref{sec:analysis} into per-label and per-change-type breakdowns.

\section{Detailed Case Studies}
\label{app:case_study}

Section~\ref{sec:analysis} reports that failure patterns differ by agent tier and that the gap between SR and MT@4 is associated with persistent-state dependency. This section presents three case studies using released task artifacts to illustrate these observations concretely. The examples address three audit questions: what makes a late round depend on earlier rounds, why SR differs from MT on the same instruction, and what information fail-stop scoring preserves.

\subsection{Case Study 1: Late-Round Conflict in a Released 15-Round Task}

\paragraph{Task and audit question.}
The released task \texttt{deterministic-data-pipeline-go} asks the agent to build \texttt{dpipe}, a deterministic Go CLI for data-pipeline reproducibility. Table~\ref{tab:dpipe_example} gives an abbreviated round chain. The main audit question is whether the final rounds are independent feature additions. They are not: later rounds revise checksum formats, lineage semantics, normalization defaults, snapshot hashes, and transforms while still requiring the earlier deterministic pipeline, validation, profiling, audit, schema, fingerprint, snapshot, and changelog behavior. Table~\ref{tab:dpipe_active_contracts} traces the active requirements through selected rounds.

\begin{table}[htbp]
\caption{Active requirements in \texttt{deterministic-data-pipeline-go}. The final round is local in wording but global in verification: the cumulative verifier tests all active requirements, not only the latest addition.}
\label{tab:dpipe_active_contracts}
\centering
\small
\begin{tabular}{cL{4.1cm}L{7.5cm}}
\toprule
\textbf{Round} & \textbf{Active requirement added or revised} & \textbf{Why it still matters later} \\
\midrule
1 & Deterministic CSV pipeline and schema validation & All later commands operate on the same deterministic data model and must preserve bit-identical output behavior. \\
3 & Boundary fill and z-score corrections; profiling and drift & Later transforms are checked against corrected numerical semantics, not only against newly added commands. \\
5 & Replace SHA-256 sidecars with BLAKE2b-256 manifest checksums; add lineage & Later manifest, reconcile, and audit behavior depends on this checksum migration and on lineage records. \\
7 & Correct lineage graph semantics; add temporal transform & Directory scans must not become false input nodes, and later lineage checks inherit this rule. \\
10 & Replace manifest field \texttt{blake2b} with \texttt{checksum} and \texttt{algorithm}; add reconcile & A correct later state must stop emitting the obsolete field while preserving manifest comparability. \\
12 & Correct default \texttt{normalize} behavior; add numeric rounding & Later transform tests use the corrected default and identical-value behavior as part of the accumulated requirements. \\
15 & Change snapshot row hashes from MD5 to SHA-256; add constant-value tags & The final verifier checks the new snapshot hash behavior together with the accumulated transforms, lineage, manifest, audit, and changelog behavior. \\
\bottomrule
\end{tabular}
\end{table}

\paragraph{Why the final round is not a local edit.}
Round 15 asks for a snapshot hash change and a tag transform. A solution that only edits the latest snapshot routine can still fail if it leaves the round-10 manifest schema inconsistent, restores obsolete SHA-256 sidecars from round 5, forgets the round-7 lineage correction, or regresses the round-12 normalization rule. The cumulative verifier operationalizes the claim behind MT@4: an agent must maintain an evolving workspace whose active requirements are distributed across earlier turns. This cross-round dependency is what makes persistent multi-turn evaluation distinct from a sequence of independent coding prompts.

\subsection{Case Study 2: Why SR Does Not Imply MT Reliability}

\paragraph{Controlled comparison.}
SR and MT@4 evaluate the same round instructions under different workspace conditions. In SR, Harbor fast-forwards the workspace with reference deltas before handing the target round to the evaluated model. In MT@4, the model receives the next instruction in the workspace it produced itself. The difference is visible on tasks such as \texttt{deterministic-data-pipeline-go}: a round-10 single-round attempt starts from a reference implementation that already satisfies the BLAKE2b migration and lineage correction, while a multi-round attempt must carry those prerequisites forward from its own earlier edits. This controlled comparison isolates the effect of accumulated workspace state on task success.

\paragraph{Concrete failure surface.}
The round-10 request replaces the manifest field \texttt{blake2b} with the pair \texttt{checksum} and \texttt{algorithm}. In SR, the model can focus on the schema rename and the new reconcile command, because the reference workspace already contains a correct BLAKE2b manifest implementation. In MT@4, the same edit is entangled with whether the model's prior implementation already removed SHA-256 sidecars, whether manifest generation is deterministic, and whether lineage entries remain consistent. SR measures isolated instruction-following skill; MT@4 measures whether earlier work remains a usable foundation for later work.

\paragraph{Metric implication.}
This distinction explains why Table~\ref{tab:main_results} reports both MT@4 and SR. A high SR with a much lower MT@4 is not contradictory: it means that the agent can often solve individual rounds from a reference state while losing reliability when later rounds depend on its own accumulated workspace. Agents with a small SR--MT@4 gap and high SR are those whose persistent workspaces closely approximate reference-quality states. This gap is the empirical basis for the claim in Section~\ref{sec:main_results} that persistent execution amplifies the difference across agent tiers.

\subsection{Case Study 3: What Fail-Stop Localizes}

\paragraph{Audit question.}
Fail-stop scoring is useful only if the failed round provides interpretable evidence. In \MYBENCH, the failed round is the first round where the agent-produced workspace no longer satisfies the cumulative verifier. Annotators then inspect which assertions failed and whether they correspond to newly introduced behavior (missed active requirement), still-active prior behavior (regression), or superseded behavior that should have been removed (stale conflict). Table~\ref{tab:case_failure_labels} illustrates each label with concrete examples from the data-pipeline task.

\begin{table}[htbp]
\caption{How fail-stop traces are interpreted during failure annotation. Each label corresponds to a distinct failure mechanism visible in the verifier output.}
\label{tab:case_failure_labels}
\centering
\small
\begin{tabular}{L{3.8cm}L{8.2cm}}
\toprule
\textbf{Failure label} & \textbf{Example evidence in the data-pipeline task} \\
\midrule
Regression introduction & A later manifest or snapshot edit breaks deterministic output, schema validation, profiling, or audit behavior that had already been required. \\
Conflict mishandling & The workspace keeps an obsolete field such as \texttt{blake2b}, keeps writing removed checksum sidecars, or continues using MD5 row hashes after SHA-256 becomes the active requirement. \\
Context loss & Corrected details such as default \texttt{normalize} behavior or lineage graph rules disappear from later edits. \\
Environment mismanagement & The Go CLI no longer builds, generated artifacts become stale, or command wiring changes in a way that prevents cumulative tests from exercising the tool. \\
\bottomrule
\end{tabular}
\end{table}

\paragraph{Why the localization matters.}
Continuing after a failure would produce additional observations, but those observations would run on a workspace that is already known to violate active requirements. The resulting data conflates two questions: whether the agent can solve the next instruction, and whether it can repair accumulated damage. Fail-stop instead records the first requirement failure and leaves later-round ability to SR or reference-fast-forward analyses. This separation is why the paper reports MT@4 for persistent reliability, SR for isolated round skill, Comp for full-task completion, and round-$k$ pass rates for where failures appear along the horizon. The diagnostic value of each metric depends on the others: MT@4 alone does not reveal whether failures cluster early or late, while round-$k$ pass rates alone do not reveal whether failures reflect persistent-state issues or instruction difficulty.

\section{Evaluation Infrastructure Details}
\label{app:infrastructure}

Section~\ref{sec:eval_protocol} describes the evaluation protocol at a design level. This appendix provides the infrastructure details: the agent harness, model configuration, Harbor extensions, reproducibility protocol, and alternative scoring designs that were considered but not adopted.

\subsection{Terminus-2 Agent Harness}
\label{app:terminus2}

Terminus-2~\cite{laude2026terminus2} is a lightweight agent harness that connects language models to the evaluation container through a uniform interface. It provides each model with shell access to the Docker container, supporting file reads and writes, command execution, package installation, and test execution. At each round, Terminus-2 delivers the round's instruction to the model, relays the model's tool calls to the container, and collects the model's completion signal. The harness imposes no constraints on the model's task-solving strategy: it may read code, run tests, install dependencies, or execute arbitrary shell commands. All models receive the same system prompt and tool definitions, reducing harness configuration as a source of variation.

\paragraph{Interaction and token accounting.}
\textbf{Avg. Turns} estimates main-agent Terminus-2 episodes per full multi-round agent-task evaluation. For each recorded trajectory, the summarizer sums \texttt{agent\_result.metadata.n\_episodes} over fragments, divides by the number of covered rounds, and multiplies by the task's total number of rounds; complete executions use the observed episode count directly. If \texttt{n\_episodes} is absent, the summarizer falls back to the number of recorded request-time entries. \textbf{Output Tok. (K)} is computed by summing \texttt{agent\_result.n\_output\_tokens} over recorded fragments, averaging trajectory totals, and dividing by 1000. Terminus-2 obtains token usage from LiteLLM response metadata. Chat-completion calls provide prompt and completion token counts; Responses API calls provide input and output token counts. Harbor accumulates input tokens, output tokens, cache-hit input tokens, and cost over the agent run. The paper reports output tokens only. This value is provider-reported generated-token usage rather than a tokenizer pass over saved transcripts; hidden reasoning tokens are included only when the provider includes them in the reported completion tokens. Harbor does not separately add provider-specific reasoning-token detail fields to \texttt{n\_output\_tokens}.

\subsection{Model Configuration}
\label{app:model_config}

All 13 models are accessed through API-compatible endpoints configured in the evaluation model configuration file. No model receives a custom task prompt beyond the standard Terminus-2 harness prompt, and no model-specific prompt engineering is applied. The main text, tables, and figures use compact labels to fit dense results; Table~\ref{tab:model_name_mapping} gives the full evaluated names and endpoint identifiers.

\begin{table}[htbp]
\caption{Mapping from compact labels used in the paper to the full evaluated model configurations. All models use the same Terminus-2 harness and Harbor evaluation protocol.}
\label{tab:model_name_mapping}
\centering
\scriptsize
\begin{tabularx}{\textwidth}{p{1.7cm}p{3.4cm}p{4.8cm}X}
\toprule
\textbf{Paper label} & \textbf{Full evaluated name} & \textbf{Endpoint model string} & \textbf{Evaluation setting} \\
\midrule
Opus-4.7 & Claude-Opus-4.7-High & \texttt{openai/claude-opus-4-7} & high reasoning effort \\
GPT-5.5 & GPT-5.5-High & \texttt{openai/gpt-5.5} & high reasoning effort \\
Opus-4.6 & Claude-Opus-4.6 & \texttt{openai/claude-opus-4-6} & default configured reasoning \\
GLM-5.1 & GLM-5.1 & \texttt{openai/z-ai/glm-5.1} & thinking enabled \\
Kimi-K2.6 & Kimi-K2.6 & \texttt{openai/moonshotai/kimi-k2.6} & thinking enabled \\
DS-V4-Pro & DeepSeek-V4-Pro & \texttt{openai/deepseek/deepseek-v4-pro} & high reasoning effort \\
Qwen3.6-Plus & Qwen3.6-Plus & \texttt{openai/qwen/qwen3.6-plus} & thinking enabled \\
MiMo-V2.5 & Xiaomi-MiMo-V2.5-Pro & \texttt{openrouter/xiaomi/mimo-v2.5-pro} & high reasoning effort \\
Gemini-3.1 & Gemini-3.1-Pro-Preview & \texttt{openai/gemini-3.1-pro-preview} & high reasoning effort \\
DS-V4-Flash & DeepSeek-V4-Flash & \texttt{openai/deepseek/deepseek-v4-flash} & high reasoning effort \\
Qwen3.5-397B & Qwen3.5-397B-A17B & \texttt{openai/qwen3.5-397b-a17b} & thinking enabled \\
MiniMax-M2.7 & MiniMax-M2.7 & \texttt{openai/minimax/minimax-m2.7} & reasoning split enabled \\
Doubao-2.0 & Doubao-Seed-2.0-Pro & \texttt{openai/doubao-seed-2.0-pro} & high reasoning effort \\
\bottomrule
\end{tabularx}
\end{table}

\subsection{Harbor Multi-Turn Extensions}
\label{app:harbor_extensions}

The Harbor framework~\cite{laude2026harbor} was originally designed around a single instruction, a single agent execution, and a single verification step. \MYBENCH~ requires different execution semantics: the same workspace and agent session must persist across rounds, while the verifier changes at each round and continues to check all still-valid requirements. We extend Harbor at the level of task semantics and trial state. The extensions are summarized below; each addresses one aspect of the persistent multi-turn protocol.

\paragraph{Round as the unit of instruction and verification.}
A multi-round task declares \texttt{num\_rounds} and one metadata record per round. Each round has its own instruction, reference delta, test script, and change-type labels. Harbor validates that the declared rounds are contiguous and that every round contains the required files. This keeps the task format explicit: there is no hidden convention that infers rounds from file names alone.

\paragraph{Persistent execution with explicit round boundaries.}
The agent works in one Docker container and one agent session across the whole task. At each boundary, Harbor delivers the next instruction, waits for the agent to finish, replaces the verifier payload with that round's cumulative tests, records the binary reward, and then either advances or stops. The agent receives a sequence of user requests; the evaluation system owns the round boundary, verifier swap, reward recording, and fail-stop decision.

\paragraph{Cumulative tests and fail-stop aggregation.}
Harbor runs the cumulative verifier for round $i$ from \texttt{round\_i/tests/}, checking all requirements from rounds 1 through $i$ that remain active. The trial result stores both per-round rewards and the aggregate window used for scoring. Missing rewards inside the scoring window count as zero, which gives the fail-stop score $S=\frac{1}{N}\sum_i r_i$ while preserving the detailed round log for analysis.

\paragraph{Reference fast-forward for controlled comparisons.}
Some analyses require testing a target round from a known-correct state. Harbor supports this by applying the reference deltas for rounds before the target round, then handing the workspace to the evaluated agent. This protocol is used for single-round evaluation (SR). It is not used for the main multi-turn score, because the main score must depend on the agent's own earlier work.

\paragraph{Resume and state lineage.}
Long multi-round evaluations need recovery from infrastructure interruption without changing the evaluation semantics. Harbor records round-boundary state snapshots, agent-session snapshots, per-round verifier outputs, and lineage metadata linking a resumed trial to its source trial and completed round. Resume is constrained by agent identity and task checksum checks, so a resumed run is a continuation of a documented prior state rather than a new evaluation condition.

\paragraph{Separation of execution and scoring windows.}
The execution window determines which rounds are actually run. The scoring window determines which round rewards contribute to the aggregate score. Keeping these windows separate supports late-round debugging, round-isolated evaluation, and ablation studies while leaving the main benchmark definition fixed.

\subsection{Reproducibility Protocol}

Each evaluation run produces a structured output directory that records all information needed to reproduce and audit the result:

\begin{verbatim}
trial/
  config.json         # Complete run configuration
  round_1/
    reward.json       # Binary reward and verification log
    state/checkpoint/ # Optional container checkpoint
    trajectory.jsonl  # Agent actions and observations
  round_2/
    ...
  round_N/
    ...
  aggregate.json      # Composite score and scoring window
\end{verbatim}

The \texttt{config.json} records the model identifier, agent harness version, Docker image hash, random seeds, and all evaluation parameters. The \texttt{aggregate.json} records the scoring window boundaries and the composite score formula, so any reported result can be reproduced exactly. The \texttt{trajectory.jsonl} files contain the full sequence of agent actions, tool calls, and observations, supporting post-hoc analysis of agent behavior at any round.

\subsection{Scoring Alternatives}
\label{app:scoring_alternatives}

Section~\ref{sec:task_def} describes fail-stop scoring as the primary mechanism and explains why it is appropriate for evaluating persistent multi-turn reliability. Two alternative designs were considered during benchmark development; we describe them here along with the reasoning behind their rejection.

\paragraph{Continue-after-failure.}
Under this design, the agent continues to receive instructions after a failed round. This approach would yield more data points per task but introduces a confound: later rounds execute on a workspace that already violates active requirements, so their results reflect the interaction between the current instruction and the accumulated damage from prior failures rather than the agent's ability on that instruction in isolation. In multi-turn coding, a broken build, a corrupted schema, or a missing dependency can cascade through later rounds in ways that obscure the failure's root cause. We considered this approach unsuitable for diagnostic evaluation, though it may be useful for studying recovery capabilities in future work.

\paragraph{Reference fast-forward.}
When the agent fails round $k$, the evaluation framework applies the reference solution for round $k$, restoring the workspace to a known-good state before proceeding to round $k+1$. This approach isolates each round's difficulty but removes the defining property of multi-turn evaluation: the dependence of later rounds on the agent's own prior work. It also inflates scores by giving the agent a foundation it did not build. We use reference fast-forward only for the SR metric, not for the primary MT@4 score.

The fail-stop model is a conservative choice that may underestimate agents capable of recovering from errors. However, it provides a clear diagnostic signal: the first failed round is the first point where the accumulated workspace no longer satisfies the cumulative verifier. The round-$k$ pass rate curves reported in Section~\ref{sec:main_results} complement the aggregate MT@4 score by revealing the shape of performance degradation, and the SR metric provides a complementary view of isolated round difficulty.

\section{Supplementary Results and Diagnostics}
\label{app:full_results}

Section~\ref{sec:main_results} reports aggregate results across all 13 agents, and Section~\ref{sec:analysis} presents fine-grained failure diagnostics. This appendix provides supplementary breakdowns that support and extend those findings. The released evaluation package includes the full per-task, per-round, and per-category data in machine-readable JSON and CSV formats.

\subsection{Per-Task Scores}

Table~\ref{tab:main_results} in the main text reports aggregate MT@4, SR, and Comp across all 26 tasks. The released evaluation package provides the corresponding per-task breakdowns: for each agent-task pair, the package includes the per-attempt round-level rewards, the MT@4 score, the single-round pass rates, agent-interaction turns, and output-token usage. These per-task scores reveal substantial variance across tasks even for the strongest agents. For example, Opus-4.7 achieves MT@4 above 80 on several short construction tasks while scoring below 30 on long migration tasks, consistent with the finding in Section~\ref{sec:analysis} that task characteristics modulate difficulty beyond aggregate averages.

\subsection{Per-Round Pass Rates}

Section~\ref{sec:main_results} reports that round-level MT@4 pass rates decline sharply within the first few rounds. Table~\ref{tab:round_pass_detail} provides the full round-by-round breakdown. Pass rates drop from 46.7\% at round 1 (all 26 tasks active) to 36.4\% at round 2, 26.9\% at round 3, 22.8\% at round 4, and 21.3\% at round 5. The decline continues more gradually through rounds 6--9 (18.8\%, 17.2\%, 15.8\%, 17.3\%). Rounds 10--15 involve progressively fewer tasks (5 tasks reach round 10, 2 tasks reach round 15), so pass rates at these horizons have higher variance and should be interpreted with caution. The fluctuations at rounds 14--15 (11.5\%) reflect the small number of tasks rather than a genuine recovery.

\begin{table}[htbp]
\caption{Round-level MT@4 pass rates averaged across all 13 agents. The number of active tasks decreases at longer horizons because tasks have different lengths (5--15 rounds).}
\label{tab:round_pass_detail}
\centering
\small
\begin{tabular}{ccc}
\toprule
\textbf{Round} & \textbf{Mean pass rate (\%)} & \textbf{Active tasks} \\
\midrule
1 & 46.7 & 26 \\
2 & 36.4 & 26 \\
3 & 26.9 & 26 \\
4 & 22.8 & 26 \\
5 & 21.3 & 26 \\
6 & 18.8 & 25 \\
7 & 17.2 & 25 \\
8 & 15.8 & 17 \\
9 & 17.3 & 12 \\
10 & 7.7 & 5 \\
11 & 7.7 & 5 \\
12 & 9.6 & 4 \\
13 & 9.6 & 2 \\
14 & 11.5 & 2 \\
15 & 11.5 & 2 \\
\bottomrule
\end{tabular}
\end{table}

\subsection{Cost and Horizon Diagnostics}
\label{app:cost_horizon}

\begin{figure*}[t]
\centering
\includegraphics[width=0.82\textwidth]{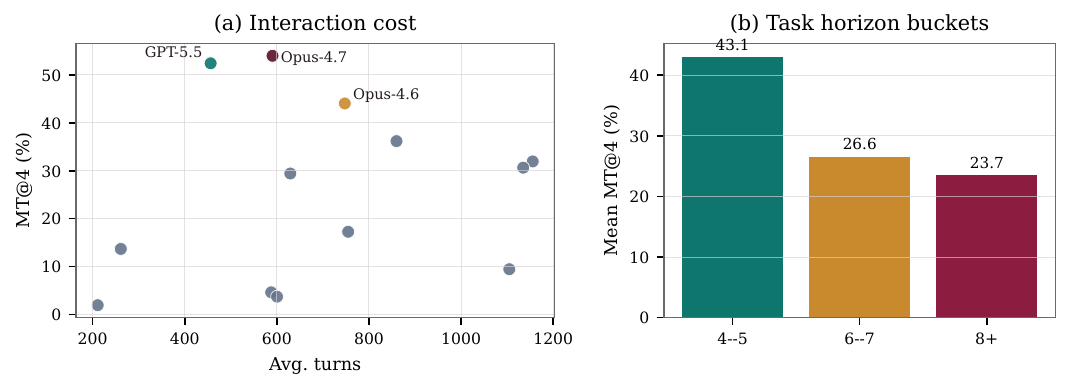}
\caption{Auxiliary diagnostics for multi-round evaluation. (a)~Average interaction turns are not monotonic with MT@4, indicating that more interaction turns do not by themselves explain persistent reliability. (b)~Mean MT@4 decreases across longer task-horizon buckets.}
\label{fig:appendix_cost_horizon}
\end{figure*}

Section~\ref{sec:main_results} notes that performance degrades across rounds and that task length is associated with lower MT@4 scores. Figure~\ref{fig:appendix_cost_horizon} provides two diagnostic plots. Panel (a) shows that average interaction turns (a proxy for computational cost) are not monotonically associated with MT@4: some low-scoring agents produce many interaction turns without improving outcomes, while some high-scoring agents are relatively efficient. Panel (b) shows that mean MT@4 decreases across task-length buckets, from shorter tasks to longer ones, consistent with accumulated workspace inconsistency contributing to the decline, though task content also co-varies with length. However, because shorter and longer tasks differ in content as well as length, this association should be interpreted as diagnostic rather than causal.

\subsection{Scores by Technology Domain}

\MYBENCH~ spans six broad technology domains: ML and MLOps (9 tasks), data engineering (4 tasks), systems and code (3 tasks), scientific computing (3 tasks), testing and automation (4 tasks), and infrastructure and security (3 tasks). Domain labels are recorded in \texttt{task.toml} and are visualized in Figure~\ref{fig:stats}. We intentionally treat domain labels as secondary to the two-dimensional taxonomy (Section~\ref{sec:task_def}), because technical domain and multi-turn behavior are not the same property: an MLOps task can be construction-oriented, review-driven, or migration-oriented depending on the requirement chain. Domain coverage instead supports external validity by ensuring that the benchmark exercises a range of programming languages, frameworks, and system types rather than concentrating on a single technology stack.

\subsection{Scores by Change Type Composition}

Section~\ref{sec:dataset_stats} reports that 110 distinct rounds carry at least one correction or conflict annotation. The failure annotation data (Appendix~\ref{app:failure_annotation}) reveals how failure patterns vary across change types. On extension rounds, the dominant failure mode is missed active requirement (90.1\% of annotated failures), with environment/tooling failures accounting for 6.5\% and regression for 2.4\%. On correction rounds, missed active requirement remains dominant (84.8\%) but regression rises to 11.2\%, consistent with the difficulty of updating previously working behavior. On conflict rounds, the failure distribution is most diverse: missed active requirement drops to 72.3\%, conflict mishandling (stale superseded behavior) rises to 14.9\%, and both environment/tooling and regression contribute 6.4\% each. This pattern supports the interpretation in Section~\ref{sec:analysis} that conflict rounds expose a qualitatively different failure surface from extension rounds.

\subsection{Scores by Initial Workspace State}

Tasks in \MYBENCH~ begin from three workspace conditions: empty or lightly scaffolded projects (construction and specification-evolution tasks), documentation-driven projects with persistent specification files (document-driven tasks), and existing legacy codebases requiring adaptation (migration tasks). The initial workspace state is part of the task design record in \texttt{task.toml}. Migration tasks are generally associated with lower MT@4 scores, consistent with the finding that adapting an existing codebase while preserving backward compatibility is more challenging than building from scratch. However, because initial workspace state co-varies with engineering activity and task content, we treat this observation as a task-design characteristic rather than an independent experimental variable.

\subsection{Impact of AGENTS.md}

Section~\ref{sec:analysis} reports that document-driven tasks average 50.1 MT@4 across agents, roughly 2.4$\times$ the mean of explorative and contractual tasks. Four of the 26 tasks use \texttt{AGENTS.md} or equivalent specification files to encode persistent quality constraints. In these tasks, the instruction may only state that the specification has been updated, and the verifier checks whether the implementation is synchronized to the current document state. The elevated MT@4 on document-driven tasks may reflect the availability of a persistent, readable reference rather than an inherently easier task structure; however, the sample size (4 tasks, 31 rounds) is too small to draw causal conclusions. Expanding the benchmark with additional document-driven tasks across different engineering activities would help determine whether the document-driven advantage generalizes.

\subsection{Per-Round SR vs.\ MT@4 Comparison}
\label{app:sr_mt_round}

Figure~\ref{fig:sr_vs_mt_round} (main text) compares single-round pass rates (SR, single attempt from a reference-fast-forwarded workspace) against multi-round MT@4 (best of four persistent-execution attempts) at each round index. SR rates remain stable between 52\% and 57\% for rounds~3--8, indicating that isolated round difficulty does not increase much with round index. In contrast, MT@4 rates decline monotonically from 46.7\% at round~1 to 7.7\% at round~10. At round~1, MT@4 exceeds SR because MT@4 selects the best of four independent attempts (pass@4 effect) while SR uses a single attempt; this ordering reverses from round~2 onward as persistent-state degradation outweighs the retry advantage. Appendix~\ref{app:cost_horizon} provides supporting task-length diagnostics.

\subsection{Cross-Attempt Variance and Reliability}
\label{app:variance}

\begin{figure*}[t]
\centering
\includegraphics[width=0.92\textwidth]{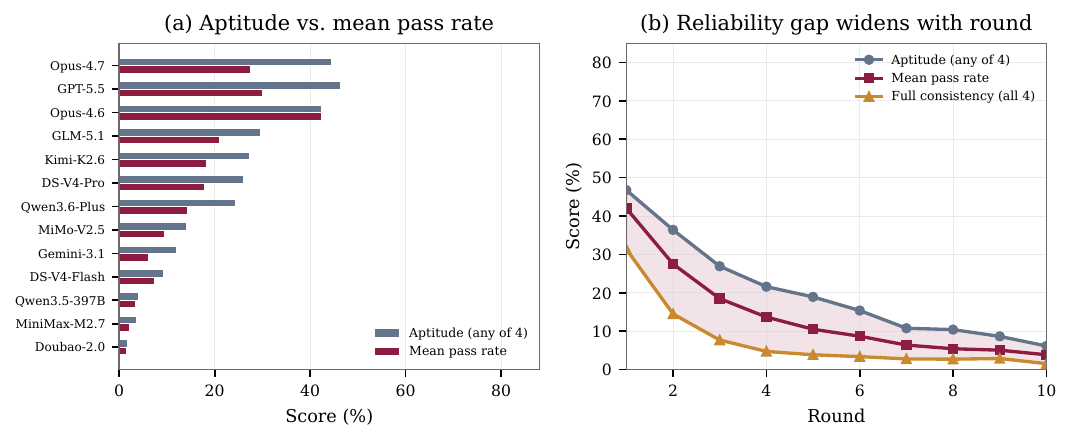}
\caption{Cross-attempt variance decomposition. (a)~Per-model aptitude (any attempt passes) versus full consistency (all attempts pass) and mean pass fraction. (b)~Per-round decomposition showing aptitude, mean pass rate, and full consistency; the shaded region between aptitude and full consistency is the reliability gap.}
\label{fig:variance_decomposition}
\end{figure*}

Figure~\ref{fig:variance_decomposition} decomposes multi-attempt performance into aptitude (at least one of four attempts passes) and full consistency (all four attempts pass). The reliability ratio (full consistency divided by aptitude) drops from 67\% at round~1 to 20\% at round~5, indicating that cross-attempt consistency declines at deeper rounds. At round~1, 31.4\% of (agent, task) pairs pass all four attempts; by round~3, only 7.7\% do, even though 26.9\% still pass at least once. This decomposition parallels the aptitude-vs-unreliability framework of \cite{bsharat2025llms} and shows that the reliability gap appears across all evaluated models: per-model breakdowns reveal that even the strongest agents exhibit declining reliability ratios at deeper rounds.

\subsection{Failure Mode Progression by Round}
\label{app:failure_progression}

\begin{figure*}[t]
\centering
\includegraphics[width=0.95\textwidth]{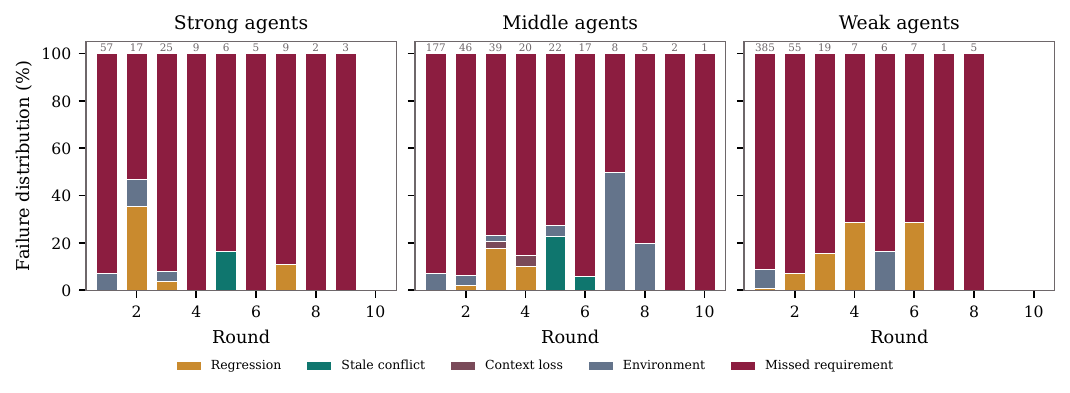}
\caption{Failure mode distribution by round index for each agent tier. Bar heights show the percentage breakdown of annotated failure modes; numbers above bars indicate the failure count at that round. Regression and conflict-mishandling failures are more visible at intermediate rounds for top-tier and mid-tier agents, while lower-tier agents are dominated by missed-requirement failures throughout.}
\label{fig:failure_progression}
\end{figure*}

Figure~\ref{fig:failure_progression} shows how the distribution of annotated failure modes changes across rounds for each agent tier. Top-tier agents exhibit a regression peak at round~2 (35\% of 17 failures), coinciding with the first correction rounds that interact with initial implementations; at later rounds, missed-requirement failures dominate as the number of top-tier failures decreases. Mid-tier agents show conflict-mishandling failures emerging at round~5 (23\% of 22 failures), absent from rounds~1--4, coinciding with the first rounds where superseded requirements accumulate. Lower-tier agents fail predominantly through missed requirements at every round (>85\%), with regression appearing only sporadically (7--29\% at rounds~2--6 on very small sample sizes). The count annotations atop each bar show that most failures concentrate at rounds~1--3 across all tiers.

\subsection{Per-Round Workspace State Penalty}
\label{app:workspace_penalty}

Section~\ref{sec:main_results} reports that SR remains stable while MT@4 declines across rounds, with the gap reaching 41 points by round~8. This appendix quantifies the workspace state penalty at the level of individual rounds. SR and MT@4 evaluate the same target instructions with the same model and cumulative verifier, but differ in workspace origin and evaluation context: SR starts from a reference-completed state, whereas MT@4 depends on the agent-produced state accumulated across prior rounds. The comparison is controlled for target instruction and verifier, and it provides evidence that accumulated workspace state contributes to the multi-turn decline.

\begin{figure*}[t]
\centering
\includegraphics[width=0.92\textwidth]{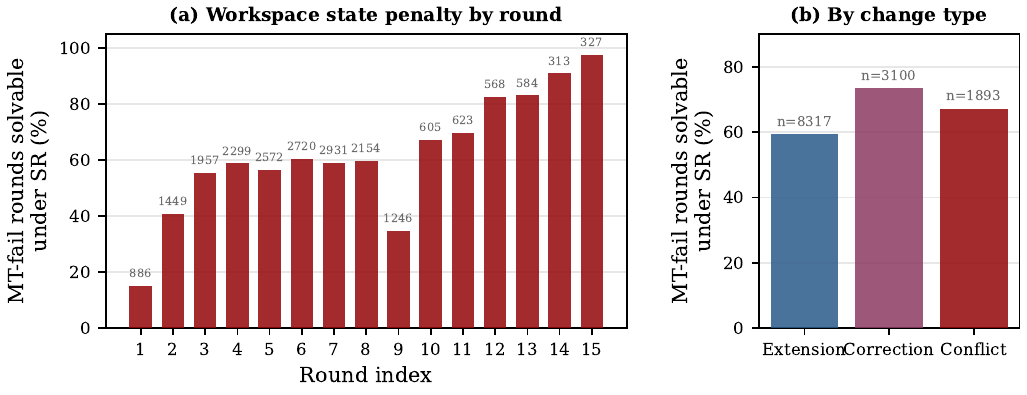}
\caption{Workspace state penalty. (a)~Percentage of MT-failing rounds that are solvable under SR, by round index. The penalty grows with depth: at later rounds, most failures involve target instructions that can be solved from a reference-completed state. Numbers above bars indicate the count of MT-failing rounds at that index. (b)~The same metric grouped by change type; correction and conflict rounds show higher SR-solvability than extension rounds.}
\label{fig:workspace_state_penalty}
\end{figure*}

Across all models and execution records, 57.0\% of individual rounds that fail under MT@4 are solvable under SR from a reference-completed state ($n{=}21{,}234$ MT-failing rounds total, of which $12{,}111$ pass under SR). The penalty grows with round depth: only 15.0\% of round-1 MT failures are SR-solvable ($n{=}886$), rising to 55.6\% at round~3 ($n{=}1{,}957$), 59.0\% at round~7 ($n{=}2{,}931$), and above 80\% beyond round~12 ($n \leq 584$). Figure~\ref{fig:workspace_state_penalty} visualizes this progression.

The penalty also varies by change type. Correction rounds show the highest fraction of SR-solvable MT failures (73.5\%, $n{=}3{,}100$), followed by conflict (67.1\%, $n{=}1{,}893$) and extension (59.4\%, $n{=}8{,}317$). This ordering is consistent with the interpretation that requirement revision is particularly sensitive to workspace quality: correction and conflict rounds demand that the agent update previously implemented behavior, which is more likely to fail when the prior workspace already deviates from a reference-quality state. The extension--correction--conflict ordering should be interpreted with caution because change types are not randomly assigned to rounds and co-vary with round position and task content.

\subsection{First-Failure-Round Distribution}
\label{app:first_fail}

Section~\ref{sec:main_results} reports that lower-tier agents fail earlier than top-tier agents. This subsection presents the full distribution. For each trial, we record the round index at which the first failure occurs and normalize by the task's total number of rounds to produce a position in $[0, 1]$. Trials that pass all rounds are excluded.

Among lower-tier agents, 57.4\% of trial failures occur in the first 20\% of rounds ($n{=}827$ failing trials), consistent with basic requirement-satisfaction deficits. Mid-tier agents show 29.4\% of failures in the first 20\% ($n{=}1{,}522$), with a more uniform distribution across positions. Top-tier agents fail earliest in only 14.4\% of cases ($n{=}993$), with failures spread more evenly and 19.1\% occurring in the final 20\% of rounds, reflecting the late-emerging regression and conflict-handling failures described in Section~\ref{sec:analysis}.

\subsection{Token Consumption and Evaluation Cost}
\label{app:token_diagnostics}

\paragraph{Evaluation cost.}
The full multi-round evaluation logged 3{,}657 execution records for 13 agents on 26 tasks and consumed approximately 24.1 billion input tokens and 333 million output tokens, or 24.4 billion total reported tokens. These records include resumed or fragmented Harbor executions rather than only the 13 $\times$ 26 $\times$ 4 top-level agent-task attempts. This total does not include single-round evaluation runs.

\begin{figure*}[t]
\centering
\includegraphics[width=0.82\textwidth]{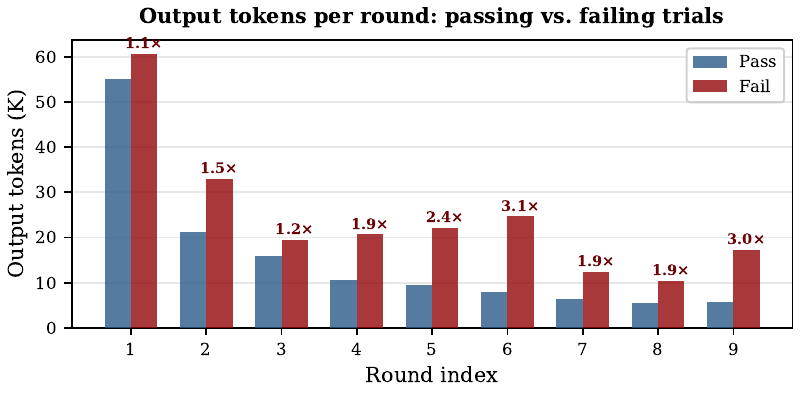}
\caption{Output tokens per round for passing vs.\ failing trials (controlled for round index). Failing trials consistently produce more output tokens at the same round position. Numbers above fail bars indicate the fail-to-pass token ratio. Sample sizes decrease at later rounds due to fail-stop termination.}
\label{fig:within_round_tokens}
\end{figure*}

\paragraph{Within-round token consumption: passing vs.\ failing trials.}
To investigate whether failing trials exhibit different resource-use patterns, we compare agent output tokens of passing and failing trials at the same round index (controlling for the confound that later rounds naturally accumulate longer contexts). Across rounds~1--9, where both pass and fail samples are sufficient, failing trials produce 1.1--3.1$\times$ as many output tokens as passing trials at the same round (Figure~\ref{fig:within_round_tokens}). The ratio is lowest at round~1 (1.1$\times$; $n_\text{pass}{=}2{,}760$, $n_\text{fail}{=}723$) and highest at round~6 (3.1$\times$; $n_\text{pass}{=}699$, $n_\text{fail}{=}54$). We emphasize that the causal direction of this association is ambiguous: more difficult workspace states may simultaneously cause both failure and increased agent effort, and the pattern should be interpreted as a diagnostic observation rather than evidence that higher token production causes failure.

\subsection{Change-Type Effect Controlling for Round Position}
\label{app:change_type_controlled}

\begin{figure*}[t]
\centering
\includegraphics[width=0.92\textwidth]{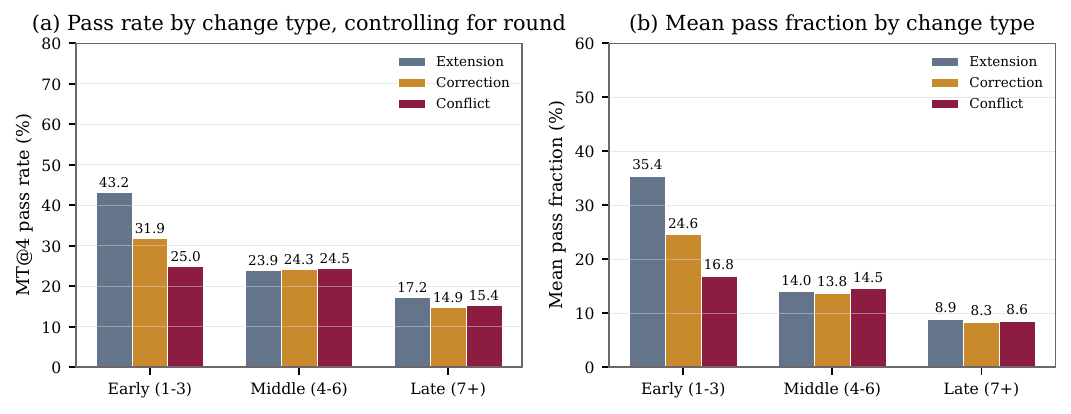}
\caption{Change-type effect on pass rates, controlling for round position. (a)~MT@4 pass rate by change type and round bucket. (b)~Mean pass fraction (across four attempts) by change type and round bucket. Extension rounds show higher pass rates than correction and conflict rounds in the early bucket, where the signal is strongest; differences diminish in the middle and late buckets.}
\label{fig:change_type_controlled}
\end{figure*}

Figure~\ref{fig:change_type_controlled} controls for round position by grouping rounds into early (1--3), middle (4--6), and late (7+) buckets, and comparing MT@4 pass rates across change types within each bucket. In early rounds, extension rounds show notably higher pass rates than correction or conflict rounds (43.2\% vs.\ 31.9\% vs.\ 25.0\%). In middle rounds, the three change types converge (23.9\% vs.\ 24.3\% vs.\ 24.5\%), and in late rounds extension again leads modestly (17.2\% vs.\ 14.9\% vs.\ 15.4\%). The early-round difference is the strongest and cleanest signal, because round position is least confounded there. Mean pass fractions (panel~b) show greater separation in the same direction for early rounds. These results support the interpretation that requirement revision (correction and conflict) imposes additional difficulty in early rounds, where the confound of round position is minimal.

\section{Responsible Release and Impact}
\label{app:responsible_release}

\paragraph{Broader impact.}
\MYBENCH~ is intended to improve the measurement of coding agents under realistic iterative development conditions. The primary positive impact is more transparent evaluation: persistent workspaces, cumulative tests, and fail-stop scoring can reveal regression and specification-tracking failures that are hidden by single-turn benchmarks. This may help researchers and practitioners deploy coding agents with more realistic expectations about long-horizon reliability.

The main negative impact is that higher-scoring coding agents can also be used to automate harmful software development. The benchmark does not include tasks for malware, credential theft, exploit development, or evasion, and released tasks focus on benign engineering workflows such as data processing, testing, migration, reporting, and configuration management. We nevertheless recommend treating benchmark improvements as capability measurements rather than deployment guarantees.

\paragraph{Assets and licensing.}
The released package includes task definitions (instructions, reference solutions, cumulative tests, and Docker environments), evaluation scripts, the Harbor multi-turn extensions, and machine-readable result summaries for all 13 evaluated models. The task format is documented in Appendix~\ref{app:task_structure}, the Harbor evaluation protocol in Appendix~\ref{app:infrastructure}, and the model configuration in Appendix~\ref{app:model_config}. Existing infrastructure assets (Harbor~\cite{laude2026harbor} and the Terminus-2 harness~\cite{laude2026terminus2}) are credited in the paper. The dataset release is marked CC-BY-NC 4.0, code and task packages include their corresponding license files, and users of closed-model APIs remain responsible for complying with the corresponding provider terms.

\end{document}